\documentclass{article}

\usepackage[final]{neurips_2021}

\usepackage[utf8]{inputenc} 
\usepackage[T1]{fontenc}    
\usepackage{hyperref}       
\usepackage{url}            
\usepackage{booktabs}       
\usepackage{amsfonts}       
\usepackage{nicefrac}       
\usepackage{microtype}      
\usepackage{xcolor}         

\usepackage{amsmath}
\usepackage{mathtools}

\newcommand{\rev}[1]{\textcolor{black}{#1}}
\newcommand{\revtwo}[1]{\textcolor{black}{#1}}

\title{Inference of Affordances and Active Motor Control in Simulated Agents}

\author{%
  Fedor Scholz \\
  Neuro-Cognitive Modeling Group\\
  University of Tübingen\\
  Tübingen, Germany \\
  \texttt{fedor.scholz@uni-tuebingen.de} \\
  \And
  Christian Gumbsch \\
  Autonomous Learning Group\\
  Max Planck Institute for Intelligent Systems\\
  \& Neuro-Cognitive Modeling Group\\
  University of Tübingen\\
  Tübingen, Germany\\
  \texttt{christian.gumbsch@tuebingen.mpg.de} \\
  \AND
  Sebastian Otte \\
  Neuro-Cognitive Modeling Group\\
  University of Tübingen\\
  Tübingen, Germany \\
  \texttt{sebastian.otte@uni-tuebingen.de} \\
  \And
  Martin V. Butz \\
  Neuro-Cognitive Modeling Group\\
  University of Tübingen\\
  Tübingen, Germany \\
  \texttt{martin.butz@uni-tuebingen.de} \\
}

\begin{document}

\maketitle

\begin{abstract}
Flexible, goal-directed behavior is a fundamental aspect of human life.
Based on the free energy minimization principle, the theory of active inference formalizes the generation of such behavior from a computational neuroscience perspective. 
Based on the theory, we introduce an output-probabilistic, temporally predictive, modular artificial neural network architecture, which processes sensorimotor information, infers behavior-relevant aspects of its world, and invokes highly flexible, goal-directed behavior.
We show that our architecture, which is trained end-to-end to minimize an approximation of free energy, develops latent states that can be interpreted as affordance maps.
That is, the emerging latent states signal which actions lead to which effects dependent on the local context.
In combination with active inference, we show that flexible, goal-directed behavior can be invoked, incorporating the emerging affordance maps.
As a result, our simulated agent flexibly steers through continuous spaces, avoids collisions with obstacles, and prefers pathways that lead to the goal with high certainty. 
Additionally, we show that the learned agent is highly suitable for zero-shot generalization across environments:
After training the agent in a handful of fixed environments with obstacles and other terrains affecting its behavior, it performs similarly well in procedurally generated environments containing different amounts of obstacles and terrains of various sizes at different locations.
\end{abstract}

\section{Introduction}

We, as humans, direct our actions towards goals.
But how do we select goals and how do we reach them?
In this work we will focus on a more specific version of the latter question:
Given a goal and some information about the environment, how can suitable actions be inferred that ultimately lead to the goal with high certainty?

The free energy principle proposed in~\cite{friston_theory_2005} serves as a good starting point for an answer.
It is sometimes regarded as a ``unified theory of the brain'' \citep{friston_free-energy_2010}, because it attempts to explain a variety of brain processes such as perception, learning, and goal-directed action selection, based on a single objective: to minimize \textit{free energy}.
Free energy constitutes an upper bound on surprise, which results from interactions with the environment.
When actions are selected in this way, we also refer to it as active inference.
Active inference basically states that agents infer suitable actions by minimizing expected free energy, leading to goal-directed planning.

One limitation of active inference-based planning is computational complexity:
Optimal active inference requires an agent to predict the free energy for all possible action sequences potentially far into the future.
This soon becomes computationally intractable, which is why so far mostly simple, discrete environments with small state and action spaces have been investigated \citep{Friston:2015}.
How do biological agents, such as humans, deal with this computational explosion when planning behavior in our complex, dynamic world?
It appears that humans, and other animals, have developed a variety of inductive biases that facilitate processing high-dimensional sensorimotor information in familiar situations \citep{Butz:2008h,EventPredictiveButz2021}.
\emph{Affordances} \citep{gibson1986ecological}, for example, encode object- and situation-specific action possibilities.
By equipping an active inference agent with the tendency to infer affordances, then, inference-based planning could first focus on afforded environmental interactions, significantly alleviating the computational load when considering interaction options.

In this work, we model these conjectures by means of an output-probabilistic, temporally predictive artificial neural network architecture.
The architecture is designed to focus on local environmental properties, from which it predicts action-dependent interaction consequences via latent state encodings. 
We show that, through this processing pipeline, \emph{affordance maps} emerge, which encode behavior-relevant properties of the environment.
These affordance maps can then be employed during goal-directed planning.
Given spatially local visual information, the resulting latent affordance codes constrain the considered environmental interactions.
As a result, planning via active inference becomes more effective and enables, for example, the avoidance of uncertainty while moving towards a given goal location. 
We furthermore show that the architecture exhibits zero-shot learning abilities \citep{Eppe:2022}, directly solving related environments and tasks within.

\newcommand\position{\mathbf{p}^t}
\newcommand\dposition{\Delta \position}
\newcommand\vision{\mathbf{v}^t}
\newcommand\sensor{\mathbf{s}^t}
\newcommand\hidden{\mathbf{h}^t}
\newcommand\action{\mathbf{a}^t}
\newcommand\context{\mathbf{c}^t}

\newcommand\positionpp{\mathbf{p}^{t+1}}
\newcommand\sensorpp{\mathbf{s}^{t+1}}

\newcommand\dpositionpp{\Delta \positionpp}
\newcommand\positionpred{\mathbf{\tilde{p}}^{t+1}}
\newcommand\dpositionpred{\Delta \positionpred}
\newcommand\sensorpred{\mathbf{\tilde{s}}^{t+1}}
\newcommand\hiddenpp{\mathbf{h}^{t+1}}

\newcommand\dpositionpreddist{\mathbf{\theta}_{\Delta \mathbf{\tilde{p}}^{t+1}}}
\newcommand\dpositionpredmu{\mathbf{\mu}_{\Delta \mathbf{\tilde{p}}^{t+1}}}
\newcommand\dpositionpredsigma{\mathbf{\sigma}_{\Delta \mathbf{\tilde{p}}^{t+1}}}

\section{Foundations}

This section introduces the theoretical foundations of our work.
We first specify our problem setting and notation.
We then introduce the free energy principle and show how we can perform active inference-based goal-directed planning with two different algorithms.
Subsequently, we combine the theory of affordances with the idea of cognitive maps and arrive at the concept of affordance maps.
We propose that the incorporation of affordance maps can facilitate goal-directed planning via active inference.

\subsection{Problem formulation and notation}

We consider problems in which an agent interacts with its environment by performing actions \rev{$\mathbf{a}$} and in turn receiving sensory states $\mathbf{s}$.
The sensory states might reveal only parts of the environmental states $\vartheta$, which therefore are not directly observable, i.e., we are facing a partially observable Markov decision process \footnote{In Markov decision processes, usually the environment additionally returns a reward in each time step, which is to be maximized by the agent.
Here, we do not define a reward function but instead plan in a model-predictive, goal-directed manner.}.
In every time step $t$, an agent selects and performs an action $\action$, and receives a sensory state (often called observation) of the next time step $\sensorpp$.

\begin{figure}
    \centering
    \includegraphics{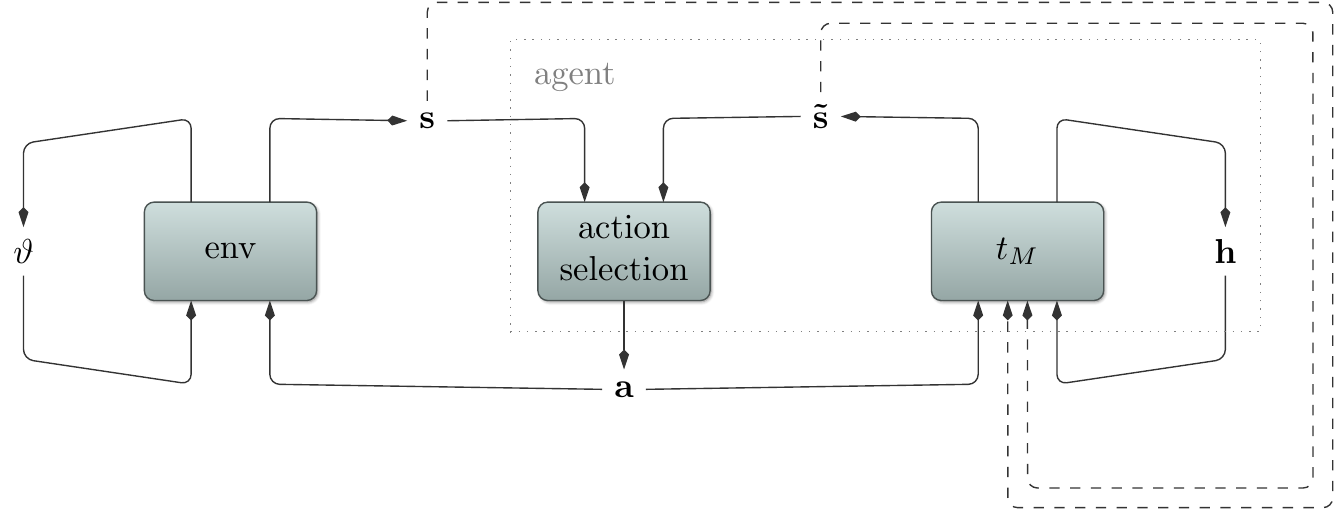}
    \caption{
    Depiction of a (partially observable) Markov decision process.
    An agent interacts with its environment by sending actions $\mathbf{a}$ and receiving consequent sensory states \rev{$\mathbf{s}$}.
    \emph{Partial observability} here implies that the sensory state \rev{$\mathbf{s}$} does not encode the whole environmental state $\vartheta$.
    Rather, certain aspects remain hidden for the agent and must be inferred from the sensory state.
    To deal with this, our agent utilizes a \rev{transition model $t_M$} with its own internal hidden state $\mathbf{h}$.
    It predicts sensory states $\mathbf{\tilde{s}}$, which aid the action selection algorithm to produce appropriate actions.
    In order to stay in tune with the environment and to predict multiple time steps into the future, the \rev{transition} model also receives observed and predicted sensory states (dashed arrows).
    }
    \label{fig:pomdp}
\end{figure}

Model-based planning, such as active inference, requires a model of the world to simulate actions and their consequences.
We use a \rev{\emph{transition model} $t_M$} that predicts the unfolding motor-activity-dependent sensory dynamics while an agent interacts with its environment.
In order to deal with partial observability, the \rev{transition} model \rev{can} be equipped with its own internal hidden state $\hidden$.
Its purpose is to encode the state of the environment, including potentially non-observable parts.
Given a current sensory state $\sensor$, an internal hidden state $\hidden$, and an action $\action$, the \rev{transition} model computes an estimate of the sensory state \rev{$\sensorpred$ in the next time step and a corresponding new hidden state $\hiddenpp$}:
\begin{equation}
    \begin{split}
        (\sensorpred, \hiddenpp) = \rev{t_M}(\sensor, \hidden, \action)
    \end{split}
    \label{eq:fm}
\end{equation}
See Figure~\ref{fig:pomdp} for a depiction of how environment, agent, \rev{transition} model, and action selection relate to each other.

\subsection{Toward\rev{s} free energy-based planning}
\label{sec:FEP}
The \emph{free energy principle} starts formalizing life itself, very generally, as having an interior and exterior, separated by some boundary \citep{Friston:2013a}.
For life to maintain homeostasis, this boundary, protecting the interior, needs to be maintained.
It follows that living things need to be in specific states because only a small number of all possible states ensure homeostasis.
The free energy principle formalizes this maintenance of homeostatic states by means of minimizing entropy.
But how can entropy be computed?
One possibility is given by the presence of an internal, generative model \rev{$m$} of the world.
In this case, we can regard entropy as the expected surprise about encountered sensory states given the model \citep{friston_free-energy_2010}.
In other words: Living things must minimize expected surprise.

This implies that all living things act as if they strive to maintain a model of \rev{their environment} over time in some way or another.
Surprise, however, is not directly accessible for a living thing.
In order to compute the surprise corresponding to some sensory input \rev{$\mathbf{s}$}, it is necessary to integrate over all possible \rev{environmental} states \rev{$\vartheta$} that could have led to that input \citep{TheFreeEnergyFristo2009}.
We can see this in the formal definition of surprise for a sensory state $\mathbf{s}$ \citep{friston_action_2010}:
\begin{equation}
    - \log p(\mathbf{s} \mid m) = - \log \int_\vartheta p(\mathbf{s}, \vartheta \mid m) \,d \vartheta
\end{equation}
where $m$ is the model or the living thing itself, and $\vartheta$ are all environmental states, including states that are not fully observable for the living thing.
The consideration of all these states is infeasible.
Thus, according to the free energy principle, living things minimize \emph{free energy}, which is defined as follows \citep{friston_action_2010}:
\rev{\begin{equation}
    FE(\mathbf{s}, \mathbf{h}) = \underbrace{E_{q(\vartheta \mid \mathbf{h})}[- \log p(\mathbf{s}, \vartheta \mid m)]}_{\textbf{energy}} - \underbrace{E_{q(\vartheta \mid \mathbf{h})}[- \log q(\vartheta \mid \mathbf{h})]}_{\textbf{entropy}}
\end{equation}
where $E$ denotes expected value and $q$ is an approximate posterior over the external hidden state $\vartheta$ given internal hidden state $\mathbf{h}$}.
Since \rev{here} all parameters are accessible, this quantity is computable.
Rewriting it shows that free energy can be decomposed into a surprise and a divergence term:
\rev{\begin{equation}
    FE(\mathbf{s}, \mathbf{h}) = \underbrace{-\log p(\mathbf{s} \mid m)}_{\textbf{surprise}} + \underbrace{D[q(\vartheta \mid \mathbf{h}) \mid \mid p(\vartheta \mid \mathbf{s}, m)]}_{\textbf{divergence}}
    \label{eq:FE}
\end{equation}}
where $D$ denotes the Kullback-Leibler divergence.
Since the divergence cannot be less than zero, free energy is an upper bound on surprise, our original quantity of interest.

Given a generative model of the world, surprise corresponds to an unexpected, inaccurate prediction of sensory information.
In order to minimize free energy, an agent equipped with a generative world model thus has two ways to minimize the discrepancy between predicted and actually encountered sensory information:
(i) The internal world model can be adjusted to better resemble the world. 
In the short term, this relates to \emph{perception}, while in the long term, this corresponds to \emph{learning}.
(ii) The agent can manipulate the world via its \emph{actions}, such that the world better fits its internal model. 
In this case, an agent chooses actions that minimize expected free energy in the future, pursuing \emph{active inference}.

\subsubsection{Active inference} \label{sec:ai}

When the free energy principle is employed as a process theory for action selection, it is called \emph{active inference}.
The name comes from the fact that the brain actively samples the world to perform inference:
It infers actions (also called control states) that minimize expected free energy (EFE), that is, \rev{an upper bound on surprise} in anticipated future states.
This is closely related to the principle of planning as inference in the machine learning and control theory communities \citep{Botvinick:2012,Lenz:2015}.
According to~\cite{Friston:2015}, a policy $\pi$ is evaluated \rev{at time step $t$} by projecting it into the future and evaluating the EFE \rev{at some time step $\tau > t$.}
Including the internal hidden states $\hidden$, EFE can be formalized as
\rev{\begin{equation} \label{eq:efe}
    EFE(\pi, t, \tau)
    = \underbrace{D[E_{p(\mathbf{h}^\tau | \hidden, \pi)} [p(\mathbf{s}^\tau \mid \mathbf{h}^\tau)] \mid \mid p(\mathbf{s}^\tau \mid m(\tau))]}_\textbf{predicted divergence from desired states}
    + \, \beta \cdot \underbrace{E_{p(\mathbf{h}^\tau | \hidden, \pi)}[H[p(\mathbf{s}^\tau \mid \mathbf{h}^\tau)]]}_\textbf{predicted uncertainty},
\end{equation}}


\rev{where $t$ is the current time step, $\tau > t$ is a future time step, and $\beta$ is a new hyperparameter that we introduce}.
This formula equates EFE with a sum of two components.
The first part is the Kullback-Leibler divergence, which estimates how far the predicted sensory states deviate from desired ones.
The second part is the entropy of the predicted sensory states, which quantifies uncertainty.
We introduce $\beta$ to weigh these components.
It enables us to tune the trade-off between choosing actions that minimize uncertainty and actions that minimize divergence from desired states.
\rev{To calculate the EFE for a whole sequence of $T$ future time steps, we take the mean of the EFE over this sequence:
\begin{equation}
    \begin{split}
        EFE(\pi, t) = \frac{1}{T} \sum_{\tau = t + 1}^{t + T} EFE(\pi, t, \tau)
    \end{split}
\end{equation}}
Based on this formula, policies can be evaluated and the policy with the least EFE can be chosen:
\begin{equation}
    \begin{split}
        \pi^t = \arg\!\min_\pi EFE(\pi, t) \\
    \end{split}
\end{equation}
Intuitively speaking, active inference-based planning agents choose actions that lead to desired sensory states with high certainty.

\subsubsection{Planning via active inference}
\label{sec:planning}
On the computational level, active inference tells us to minimize EFE to perform goal-directed planning.
Thus, it provides an objective to optimize actions.
However, it does not specify how to optimize the actions on an algorithmic level.
We thus detail two planning algorithms that can be employed for this kind of action selection.
In both algorithms, we limit ourselves to a finite prediction horizon \rev{$T$} with fixed policy lengths.
In order to evaluate policies, both algorithms employ a \rev{transition model $t_M$} and ``imagine'' the execution of a policy:
\begin{equation}
    (\mathbf{\tilde{s}}^{\tau+1}, \mathbf{h}^{\tau+1}) =
    \begin{cases}
        \rev{t_M}(\mathbf{s}^t, \mathbf{h}^t, \mathbf{a}^t), & \text{if } \tau = t \\
        \rev{t_M}(\mathbf{\tilde{s}}^\tau, \mathbf{h}^\tau, \mathbf{a}^\tau), & \text{if } t < \tau < t+T
    \end{cases}
\end{equation}
For active inference-based planning, we can compute the EFE for the predicted sequence and optimize the actions using one of the planning algorithms.
After a fixed number of optimization cycles, both algorithms return a sequence of actions. 
The first action can then be executed in the environment.  

\paragraph*{Gradient-based active inference}

Action inference \citep{lintas_inferring_2017, butz_learning_2019} is a gradient-based optimization algorithm for model-predictive control.
Therefore, it requires the \rev{transition model $t_M$} to be differentiable.
The algorithm maintains a policy \rev{$\pi$}, which, in each optimization cycle, is fed into the \rev{transition} model.
Afterwards, we use backpropagation through time to backpropagate the EFE onto the policy.
We obtain the gradient by taking the derivative of the EFE with respect to an action $\mathbf{a}^\tau$ from the policy.
After multiple optimization cycles, the algorithm returns the first action of the optimized policy.

\paragraph*{Evolutionary-based active inference}

The cross-entropy method (CEM,~\citealp{rubinstein1999cross}) is an evolutionary optimization algorithm.
CEM maintains the parameters of a probability distribution and minimizes the cross-entropy between this distribution and a distribution that minimizes the given objective.
It does so by sampling candidates, evaluating them according to EFE, and using the best performing candidates to estimate the parameters of its probability distribution for the next optimization cycle.
Recently, CEM has been used as a $\rev{zero-order}$ optimization technique for model-based control and reinforcement learning \rev{(RL)} \citep{PETS, PlaNet, iCEM}.
In such a model-predictive control setting, CEM maintains a sequence of probability distributions and candidates correspond to policies.
After multiple optimization cycles, the algorithm returns the first action of the best sampled policy.

\subsection{Behavior-oriented predictive encodings}

In theory, given a sufficiently accurate model, active inference enables an agent to plan goal-directed behavior regardless of the complexity of the problem.
In practice, however, considering all possible actions and consequences thereof quickly becomes computationally intractable. 
To counteract this problem, it appears that humans and other animal have developed a variety of inductive learning biases to focus the planning process by means of behavior-oriented, internal representations.
Here, we focus on biases that lead to the development of affordances, cognitive maps, and, in combination, affordance maps. 

\subsubsection{Affordances}

\citet{gibson1986ecological} defines \emph{affordances} as what the environment offers an animal:
Depending on the \rev{current environmental context}, affordances \rev{are} possible interactions.
As a result, affordances fundamentally determine how animals behave depending on their environment.
They constitute behavioral options from which the animal can select suitable ones in order to fulfil its current goal.
To give an example, imagine a flat surface at the height of a human's knees.
Given the structure underneath is sufficiently sturdy, it is possible to sit on the surface in a way that requires relatively little effort.
Therefore, such a surface is \emph{sit-upon-able}:
It offers a human the possibility to sit on it in an effortless way.

\rev{In this work, we use a more general definition of affordances.
We define an affordance as anything in the environment that locally influences} \revtwo{the effects of the agent's actions.
These definitions differ with respect to the set of possible actions.
Gibson's definition entails that certain actions are possible only in certain environmental contexts:
For example sitting down is only possible in the presence of a chair.
In this work, we assume that every action is possible everywhere in the environment, but that the effects differ depending on the environmental context:
Sitting down is also possible in the absence of a chair, but the effect is certainly different.
}

The theory of affordances explicitly states that to (visually) perceive the environment is to perceive what it affords.
Animals do not see the world as it is and derive their behavioral options from their perspective.
Rather, \citeauthor{gibson1986ecological} proposes that affordances are perceived directly, assigning distinct meanings to different parts of the environment.
From an ecological perspective, it appears that vision may have evolved for exactly this purpose \citep{gibson1986ecological}:
to convey what behaviors are possible in the current situation.
First, however, an animal needs to learn the relationship between visual stimuli and their meaning for behavior.
This is non-trivial:
Similar visual stimuli can mean different things, or the other way round.
Furthermore, visual input is rich such that the animal needs to effectively focus on the behavior-relevant information.

\subsubsection{Cognitive maps}

The concept of \emph{cognitive maps} was introduced in~\cite{tolman1948cognitive}.
\citeauthor{tolman1948cognitive} showed that after exploring a given maze, rats were able to navigate towards a food source regardless of their starting position.
He concluded that the rats acquired a mental representation of the maze: a cognitive map.
Place cells in the hippocampus seem to be a promising candidate for the neural correlate of this concept \citep{o1978hippocampus}.
These cells tend to fire when the animal is at associated locations.
Visual input acts as stimuli, but also the olfactory and vestibular senses play a role.
Together, place cells constitute a cognitive map, which the animal appears to use for orientation, reflection, and planning \citep{Diba:2007,Pfeiffer:2013}.

\subsubsection{Affordance maps}

Cognitive maps are well-suited for flexible navigation and goal-directed planning.
However, to improve the efficiency of the planning mechanisms, it will be useful to encode behavior-relevant aspects, such as the aforementioned affordances, within the cognitive map.
Accordingly, we combine the theory of affordances with cognitive maps, leading to \emph{affordance maps}.
Their function is to map spatial locations onto affordance codes.
Like cognitive maps, their encoding depends on visual cues.
In contrast to cognitive maps' traditional focus on map-building, though, affordance maps signal distinct behavioral options at particular environmental locations.
As an example, consider a hallway corner situation with corridors to your right and behind you.
An affordance map would encode successful navigation options for turning to the right or turning around.
\rev{Regarding affordances for spatial navigation specifically, \citeauthor{bonner2017coding} showed that these are automatically encoded by the human visual system independent of the current task and propose a location in the brain where a neural correlate could be situated \citep{bonner2017coding}.}

\subsection{Related neural network models}

\cite{1803.10122v4} used a world model to facilitate planning via \rev{RL}.
Their overall architecture consists of a vision model which compresses visual information, a memory module, and a controller model, which predicts actions given a history of the compressed visual information.
Their vision model is given by a variational autoencoder, which is trained in an unsupervised manner to reconstruct its input.
Therefore, and in contrast to ours, their vision model is not trained to extract meaningful, behavior-oriented information.
This is why we would not regard the emerging compressed codes as affordance codes.
Affordance maps were used before in~\cite{LearningToMov} to aid planning.
The authors put an agent into an environment (VizDoom) with hazardous regions that were to be avoided.
The agent moved around in its environment and collected experiences of harm or no harm, which were backprojected onto the pixels of the input to the agent's visual system, thereby performing image segmentation.
The authors then trained a convolutional neural network (CNN) on the resulting data of which the output was utilized by the A* algorithm for planning.
In contrast to ours, their architecture was not trained in an end-to-end fashion, meaning that the resulting affordance codes were not optimized to suit their \rev{transition} model.

\section{Model}

We now detail the proposed architecture, which learns a \rev{transition} model of the environment with spatial affordance encodings.
The architecture predicts a probability density function over changes in sensory states given the current sensory state and action \rev{as well as potentially an internal hidden state}.
This action-dependent \rev{transition} model of the environment thus enables active inference-based planning. 
We first specify the architecture, then detail the model learning mechanism, and finally turn to active inference.

\subsection{Affordance-conditioned inference}
\begin{figure}
    \centering
    \includegraphics[width=0.95\linewidth]{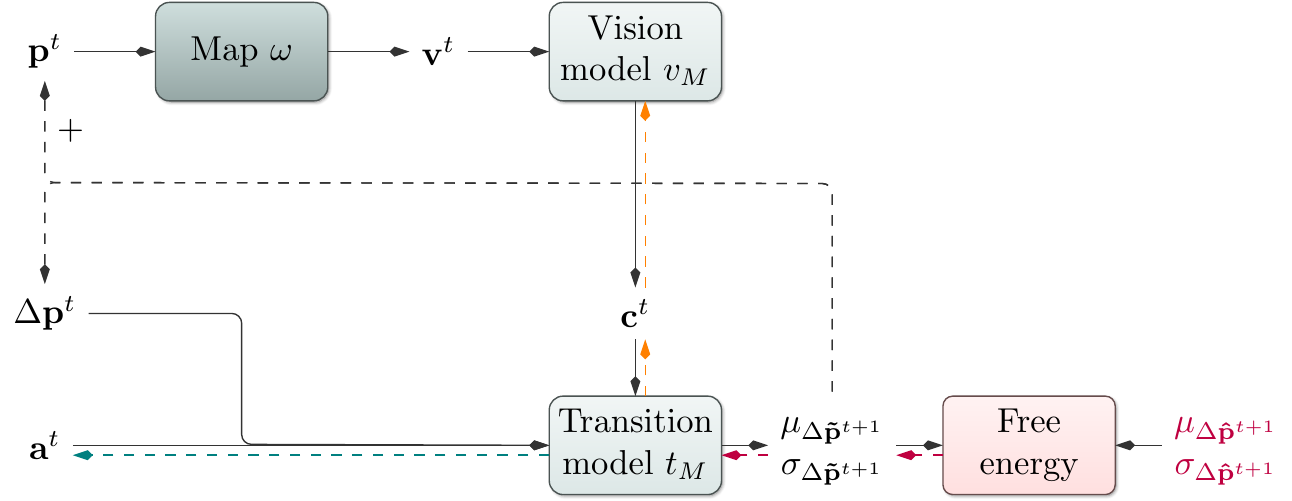}
    \caption{
    Affordance map architecture: Based on the current position $\position$, the architecture performs a look-up in an environmental map \rev{$\omega$}.
    The vision model $v_M$ receives the resulting visual information $\vision$ and produces a contextual code $\context$.
    The \rev{transition model $t_M$} utilizes this context $\context$, the last change in the position $\dposition$, an action $\action$, and its internal hidden state $\hidden$ to predict a probability distribution over the next change in position.
    During training, the loss between predicted and actual change in position is backpropagated onto \rev{$t_M$} (red arrows) and further onto $v_M$ (orange arrows) to train both models end-to-end.
    During planning, the map look-up is performed using position predictions. For gradient-based active inference, EFE is backpropagated onto the action code $\action$ (red and green arrows). 
    For planning with the cross-entropy method, $\action$ is modified directly via evolutionary optimization.
    }
    \label{fig:architecture}
\end{figure}

Our model adheres to the general notion introduced above (cf. Equation~\ref{eq:fm}).
Our model consists of three main components: a \rev{transition model $t_M$}, a vision model $v_M$, and a look-up map \rev{$\omega$} of the environment.
The model with its different components is illustrated in Figure~\ref{fig:architecture}

\rev{Our system learns a transition model $t_M$ of its environment.
It receives a current sensory state $\sensor$ and action $\action$ and predicts a consequent sensory state $\sensorpred$.
If the environment is only partially observable, the transition model can furthermore receive an internal hidden state $\hidden$ and predict a consequent $\hiddenpp$.}
Focusing on motion control tasks, we encode the \rev{sensory} state by a two-dimensional positional encoding $\position$, where the \rev{transition} model continues to predict changes in positions $\dpositionpred$ given the last positional change $\dposition$, potentially hidden state $\hidden$, and current action $\action$.
To enable the model to consider the properties of different regions in an environment during goal-directed planning, though, 
we introduce an additional contextual input $\context$, which is able to modify the \rev{transition} model\rev{'s} predictions (cf. \citealt{butz_learning_2019}, for a related approach without map-specificity).
In each time step, the \rev{transition model $t_M$} thus additionally receives a context encoding vector $\context$,
which should encode the \rev{locally} behavior-relevant characteristics of the environment.

This context code is produced by the \emph{vision model} $v_M$, which receives a visual representation $\vision$ of the agent's current surroundings in the form of a small pixel image.
The vision model is thus designed to generate vector embeddings that accurately modify the \rev{transition} model's predictions context-dependently. 


The prediction of the \rev{transition} model can thus be formalized as follows:
\begin{equation}
    \begin{split}
        (\dpositionpred, \hiddenpp) = \rev{t_M}(\context = v_M(\vision), \dposition, \hidden, \action)
    \end{split}
\end{equation}
The visual information $\vision$ can be understood as a local view of the environment surrounding the agent.
Thus, $\vision$ depends on the agent's location $\position$.
To enable the model to predict $\vision$ for various agent positions, for example, for "imagined" trajectories while planning, the system is equipped with a \emph{look-up map} \rev{$\omega$} to translate positions $\position$ into local views of the environment $\vision$.
We augment the model with the ability to probe particular map locations, translate the location into a local image, and extract behavior-relevant information from the image. 
Intuitively speaking, this is as if the network can put its focus of attention to any location on the map and consider the context-dependent behavioral consequences at the considered location.
As a result, the system is able to consider behavioral consequences dependent on probed environmental locations.
In future work, the learning of completely internal maps may be investigated further. 

The consequence of this model design is that the context code $\context$ will tend to encode \rev{local}, behavior-influencing aspects \rev{of the environment}, that is, affordances.
The context is therefore a compressed version of the environment's behaviorally-relevant characteristics at the corresponding position.
Therefore, the incorporation of the affordance codes can be expected to improve both the accuracy of action-dependent predictions and active inference-based planning.
\rev{This connection between active inference and affordances can further be described as follows \citep{friston2012dopamine}:
The desired sensory state encoded as a prior lets the agent expect to reach the target.
If the agent then is in front of an obstacle e.g., different affordances compete with each other, which is in line with the affordance competition hypothesis \citep{cisek2007cortical}:
fly around or crash into the obstacle.
Since flying around the obstacle best explains the sensory input in light of the prior, the action corresponding to this affordance is chosen by the agent.}

\subsection{Uncertainty estimation} \label{sec:active_inference}

The free energy principle is inherently probabilistic and therefore active inference requires our architecture to produce probability density functions over sensory states.
We implement this in terms of a \rev{transition model $t_M$} that does not predict a point estimate of the change in sensory state in the next time step, but rather the parameters of a probability distribution over this quantity.
We choose the multivariate normal distribution with diagonal covariance matrix (i.e., covariances are set to $0$).
The output of the \rev{transition} model is then given by a mean vector $\dpositionpredmu$ and a vector of standard deviations $\dpositionpredsigma$.
We thus replace $\dpositionpred$ with $\dpositionpreddist := (\dpositionpredmu, \dpositionpredsigma)$.

\subsection{Training}

We \rev{train both components of} our architecture \rev{jointly} in an end-to-end, self-supervised fashion to perform one-step ahead predictions \rev{on a pregenerated data set} via backpropagation through time.
The gradient flow during training is depicted in Figure~\ref{fig:architecture}.
Inputs consist of the sensor-action tuples described above.
The only induced target is given by the change in position in the next time step $\Delta \positionpp$.
This target signal is compared to the output of the \rev{transition model $t_M$} by the negative log-likelihood (NLL)\footnote{See Appendix~\ref{app:nllderivative} for a description of how to compute gradients when the objective is given by the NLL in a multivariate normal distribution.}, which approximates free energy assuming no uncertainty in our point estimate $\mathbf{h}$ of environmental state $\vartheta$ (see Equation \ref{eq:fm}).
Due to end-to-end backpropagation, the vision model \rev{$v_M$} is trained to output compact, \rev{transition} model-conditioning representations of the visual input.

While we use the NLL as the objective during training here, \rev{we make use of the expected free energy during goal-directed control}.
In future work one could utilize full-blown FE \rev{also during training} in a probabilistic architecture.
However, there is a close relationship between NLL and FE due to the Kullback-Leibler divergence:
In Appendix~\ref{app:nllferelation}, we show that minimizing NLL is equivalent to minimizing the Kullback-Leibler divergence up to a constant factor and a constant.
Thus, through NLL-based learning we can approximate learning through FE minimization.

\subsection{Goal-directed control}
We perform goal-directed control via gradient- \rev{and} evolutionary-based active inference as described in Section~\ref{sec:ai}.
Usually, in order to predict multiple time steps into the future given a policy, the \rev{transition model $t_M$} receives its own output as input.
Since our architecture predicts the parameters of a normal distribution, we use the predicted mean as input in the next prediction time step.
The model incorporates visual information \rev{$\mathbf{v}$} from locations corresponding to the predicted means.
Therefore, the model does not blindly imagine a path but simultaneously ``looks'' at, or focuses on, predicted positions, incorporating the inferred affordance code \rev{$\mathbf{c}$} into the \rev{transition} model\rev{'s} predictions.

In order to compare the predicted path to the given target and to look up the visual information, we need absolute locations.
We thus take the cumulative sum and add the current absolute position.
To compute EFE along a predicted path we also need to consider the standard deviations at every point.
For that, we first convert standard deviations to variances, compute the cumulative sum, and convert back to standard deviations.
We then can compute the EFE between the resulting sequence of probability distributions over predicted absolute positions and the given target according to Equation~\ref{eq:efe}.
To do so, we encode the target with a multivariate normal distribution as well, setting the mean to the target location and the standard deviation to a fixed value.
We can thus optimize the policy via the gradient- or evolutionary-based EFE minimization method introduced in Section~\ref{sec:planning} above\footnote{Appendix~\ref{app:planning} summarizes the particular adjustments we applied to these algorithms.}.

\section{Experiments and results}

To evaluate the abilities of our neural affordance map architecture, 
we first introduce the environmental simulator and specify our evaluation procedure in general.
The individual experimental results then evaluate the system's planning abilities to avoid obstacles and regional uncertainty as well as to generalize to unseen environments.
With respect to the affordance codes \rev{$\mathbf{c}$}, we show emerging affordance maps and examine disentanglement.

\subsection{Environment}\label{section:environment}

\begin{figure}
    \centering
    \includegraphics[width=0.5\linewidth]{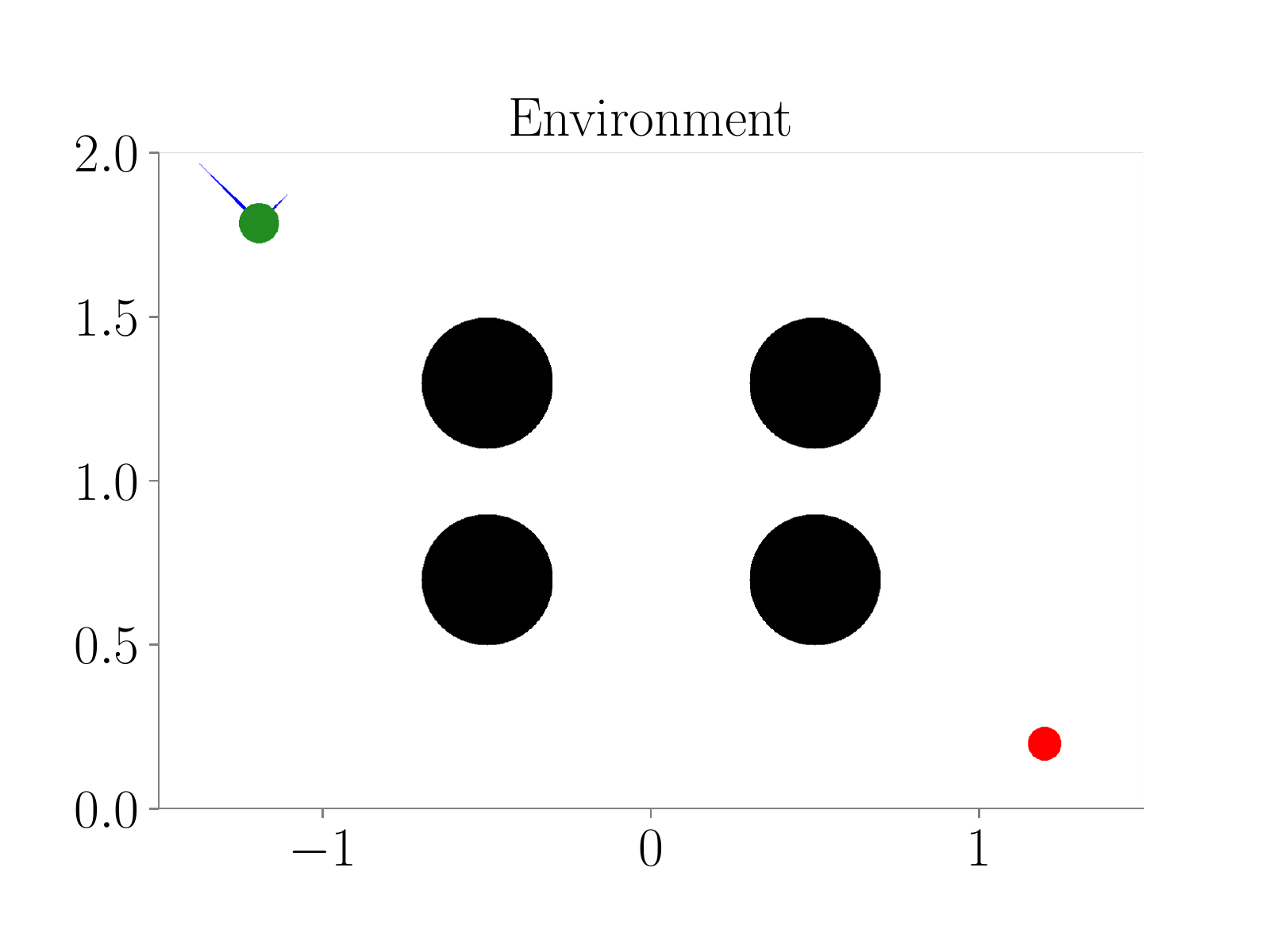}
    \caption{
        The simulation environment we use in our experiments.
        It resembles a confined space in which a vehicle (green) can move around by adjusting its throttles (blue).
        Its goal is to navigate towards the target (red).
        Depending on the experiment, obstacles or different terrains (black) are present, which affect the vehicle's sensorimotor dynamics.
    }
    \label{fig:env}
\end{figure}

The environment used in our experiments is a physics-based simulation of a circular, vehicle-like agent with radius $0.06$ in a 2-dimensional space with an arbitrary size of $3 \times 2$ units.
It is confined by borders, which prevent the vehicle from leaving the area.
The vehicle is able to fly around in the environment by adjusting its $4$ throttles, which are attached between the vertical and horizontal axes in a diagonal fashion.
They take values between $0$ and $1$ resembling actions and enable the vehicle to reach a maximum velocity of approximately $0.23$ units per time step within the environment.
Therefore, at least $13$ time steps are required for the vehicle to fly from the very left to the very right.
Due to its mass, the vehicle undergoes inertia and per default, it is not affected by gravity.
See Figure~\ref{fig:env} for a depiction of the environment and an agent.
It is implemented as an OpenAI Gym \citep{brockman_openai_2016}.

The environment can contain obstacles, which block the way.
Friction values are larger when the vehicle touches obstacles or borders.
Furthermore, the environment can comprise different terrains, which locally change the sensorimotor dynamics.
Force fields pull the agent up- or downwards.
If the vehicle is inside a fog terrain, the environment returns a position that is corrupted by Gaussian noise.
Two values from a standard normal distribution are sampled and added to each coordinate.
This implies a standard deviation of approximately $1.414$ on the difference between positions from two consecutive time steps within fog. \footnote{Since $\sqrt{1^2 + 1^2} = \sqrt{2} \approx 1.414$.}

\rev{The environment outputs absolute positions, the change in the positions and allows probing of the map at arbitrary positions.
Therefore, and apart from the noisy positions in fog terrains and the map having a lower resolution than the environment itself, the environment used in our experiments is fully observable.
This makes the incorporation of the internal hidden state $\mathbf{h}$ in the transition model $t_M$ obsolete.
In the future we plan to test our architecture on environments which are only partially observable.}

\subsection{Model and agent}

The vision model \rev{$v_M$} is given by a CNN, which produces the context activations \rev{$\mathbf{c}$}.
We always evaluate contexts of sizes $0, 1$, \rev{$3$, $5$, and $8$}.
\rev{Since the environment in our experiments is almost fully observable, we drop the internal hidden state $\mathbf{h}$ and use a multilayer perceptron (MLP) as our transition model $t_M$.}
\rev{It} consists of a \rev{fully connected layer} followed by two parallel fully connected layers for the means and standard deviations, respectively.
For each setting, we train $25$ versions of the same architecture with different initial weights \rev{on a pregenerated data set} and report \rev{aggregated} results.
See Appendix~\ref{app:model} for more details on the \rev{model and training} hyperparameters, \rev{how the visual input $\mathbf{v}$ is constructed, how the data set is generated,} and \rev{the} training procedure.

Active inference performance is evaluated after performing $100$ goal-directed control runs per setting for $200$ time steps.  
For each trained architecture instance, we consider $4$ distinct start and target positions corresponding to each corner of the environment.
The start position is chosen randomly with a uniform distribution over a $0.2 \times 0.2$ units square with a distance of $0.1$ units to the borders.
The target position is chosen in the same way in the diagonally opposite corner.
We consider the agent to have reached the target when its distance to the target falls below $0.1$ units.
The prediction horizon always has a length of $20$.
For active inference based planning (Equation~\ref{eq:efe}), the target is provided to the system as a Gaussian distribution with standard deviation $0.1$.  
We reduce the standard deviation of the target distribution to $0.01$ once the agent comes closer than $0.5$ units.
The two different standard deviations can be seen as corresponding to, for example, smelling and seeing the target, respectively.
See Appendix~\ref{app:planning} for more details on the hyperparameters.
All hyperparameters were optimized empirically or with Hyperopt \citep{bergstra2013making} via Tune \citep{liaw2018tune}.

In order to get an idea about the nature of the emerging affordance codes \rev{$\mathbf{c}$}, we plot affordance maps by generating position-dependent context activations via the vision model \rev{$v_M$} for each possible location in the environment.
That is, in the affordance maps shown below, the x- and y-axes correspond to locations in the environment while the color of each dot represents the context activation at that position.
\rev{For this, we performed a principal component analysis (PCA) on the resulting context activations in order to reduce the dimensionality to $3$ and then interpreted the results as RGB values.}

\subsection{Experiment I: Obstacle avoidance} \label{sec:avoiding_obstacles}

The first of our experiments examines our architecture's ability to avoid obstacles during active inference through the use of affordance codes.
As a baseline experiment, we consider context size $0$, disabling information flow from the vision model \rev{$v_M$} to the \rev{transition} model \rev{$t_M$}.
With context sizes larger than $0$ , however, the \rev{transition} model can be informed about obstacles and borders via the context.
 
We train the architecture on the environment depicted in Figure~\ref{fig:env}, where black areas resemble obstacles.
\rev{One-hot encoded} visual information \rev{$\mathbf{v}$} has one channel only for the obstacles and borders.
We perform goal-directed control on the same environment we train on.

\begin{figure}[htb!]
    \centering
    \includegraphics[width=0.99\linewidth]{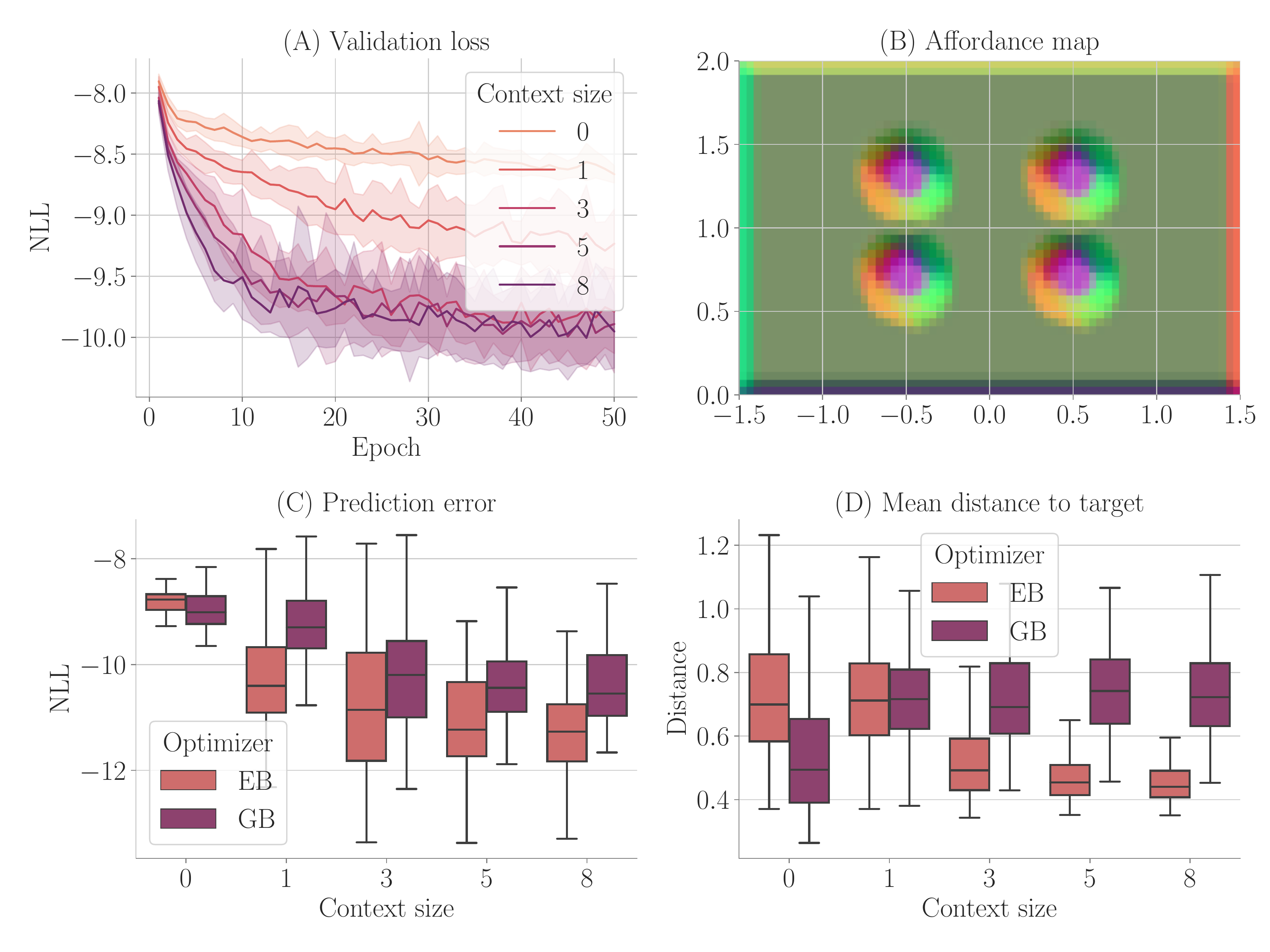}
    \caption{
        \rev{
        Results for Experiment I -- Obstacle avoidance (Subsection~\ref{sec:avoiding_obstacles}).
        For each setting, the results are aggregated over $25$ differently initialized models which each performed $4$ goal-directed control runs for $200$ time steps.
        The box plots show the medians (horizontal bars), quartiles (boxes), and minima and maxima (whiskers).
        Data points outside of the range defined by extending the quartiles by $1.5$ times the interquartile range in both directions are ignored.
        EB is short for evolutionary- and GB for gradient-based active inference.
        (A) Validation loss during training.
        It is the negative log-likelihood of the actual change in position in the transition model's predicted probability distribution.
        Shaded areas represent standard deviations.
        (B) Exemplary affordance map for context size $8$.
        To generate this map, we probed the environmental map at every sensible location, applied the vision model to each output, performed dimensionality reduction to $3$ via PCA, and interpreted the results as RGB values.
        (C) Prediction error during goal-directed control.
        It is the negative log-likelihood of the actual change in position in the transition model's predicted probability distribution.
        (D) Mean distance to the target during goal-directed control.
        }
    }
    \label{fig:exp1_results}
\end{figure}

\begin{figure}[htb!]
    \centering
    \includegraphics[width=\linewidth]{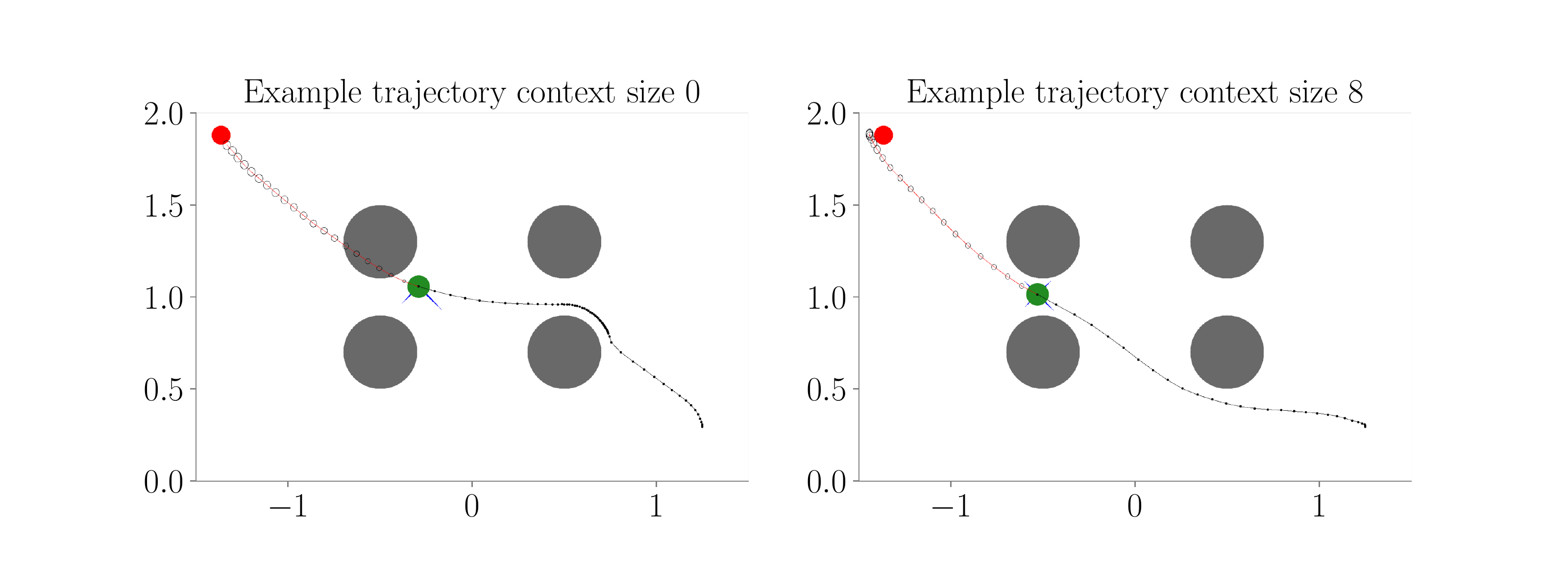}
    \caption{
        \rev{
        Example trajectories from Experiment I -- Obstacle avoidance (Subsection~\ref{sec:avoiding_obstacles} with evolutionary-based active inference.
        The agent (green) flies towards the target (red) with obstacles (grey) it its way.
        The black line behind the agent shows its past trajectory and the line in front its planned trajectory.
        Circles in front of the agent show the predicted uncertainty in the sensory states.
        With context size $0$, the agent cannot incorporate information about the environment and therefore plans thorugh and flies against the obstacles.
        With context size $8$, the agent can successfully plan its way around and therefore avoid obstacles.
        }
    }
    \label{fig:exp1_trajectories}
\end{figure}

Figure~\ref{fig:exp1_results} shows the results.
\rev{Context codes of increasing dimensionality lead} to smaller validation losses (Figure~\ref{fig:exp1_results}A), indicating their utility in improving the \rev{transition} model's accuracy.
The affordance map (Figure~\ref{fig:exp1_results}B) shows that obstacles are encoded differently from the rest of the environment.
\rev{Areas where free flight if possible are encoded with a context code that corresponds to olive green.
In contrast, areas where it is only possible to fly upwards, to the left, or downwards e.g. are encoded with a context code that corresponds to the color orange.
Light green, on the other hand, represents areas where only movement to the left is blocked.
The affordance map reveals that different sides of the obstacles are encoded similarly to the corresponding sides of the environment's boundary.
Moreover, we find gradients in the colors when moving away from borders or obstacles, indicating that the context codes not only encode directions but also distances to impassable areas.
This confirms that the emerging context codes constitute behavior-relevant encodings of the visually perceived environment.}
We evaluate goal-directed planning in terms of prediction error (Figure~\ref{fig:exp1_results}C) \rev{and} mean distance to the target (Figure~\ref{fig:exp1_results}D).
\rev{For evolutionary-based active inference, w}e find improvement \rev{in both metrics with increasing context sizes}.
\rev{For gradient-based active inference, we find improvement in the prediction error but deterioration of the mean distance to the target with increasing context sizes.}
Gradient-based outperforms evolutionary-based active inference with context size $0$.
With larger context sizes, \rev{evolutionary-based active inference performs better}.
\rev{Figure~\ref{fig:exp1_trajectories} shows two example trajectories for context sizes $0$ and $8$.
A context size of $8$ allows the agent to incorporate local information about the environment, resulting in past and planned trajectories that bend around obstacles.
With context size $0$, however, it can be seen that the agent flew against one obstacle and plans its trajectory through another one.}

\subsection{Experiment II: Generalization} \label{sec:generalization}

In this experiment we examine how well our architecture is able to generalize to similar environments.
In Experiment I (Subsection~\ref{sec:avoiding_obstacles}), we trained on a single environment.
Once the architecture is trained, we expect that our system should be able to successfully perform goal-directed control in other environments as well, given we provide the corresponding visual input.
The local view onto the map essentially allows us to change position and size of obstacles without expecting significant deterioration in performance.

\begin{figure}[htb!]
    \centering
    \includegraphics[width=\linewidth]{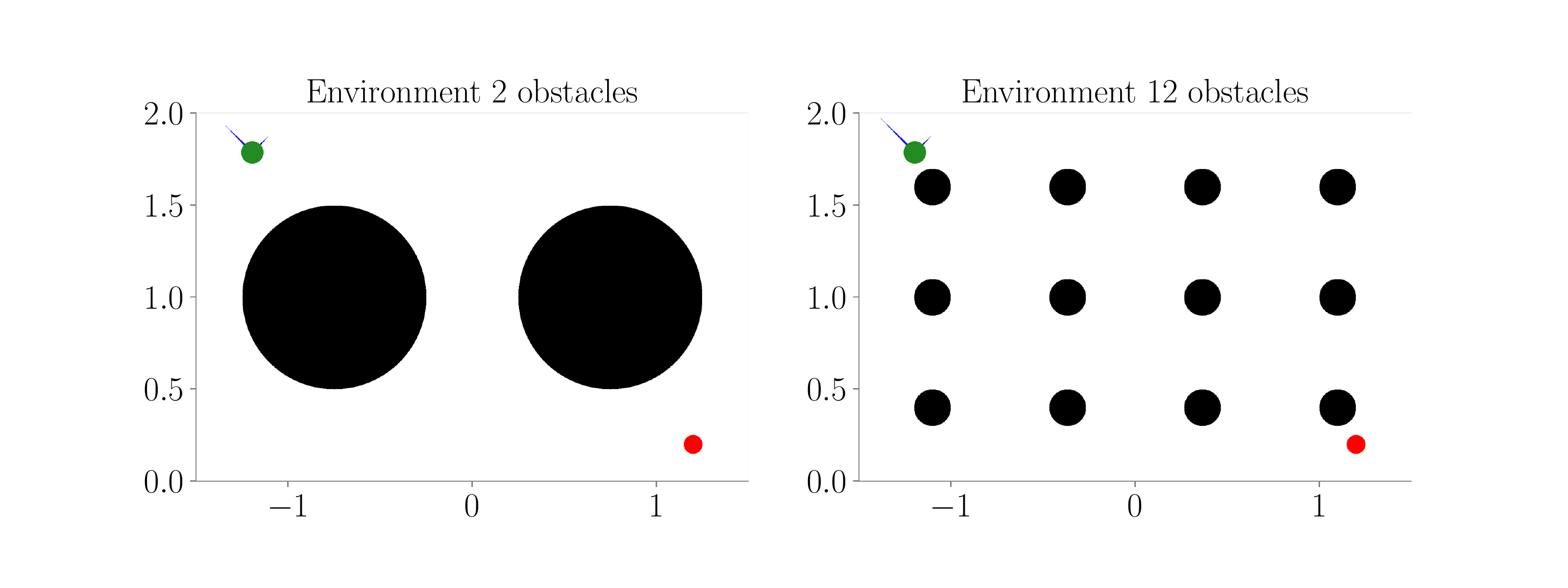}
    \caption{Additional environments used during goal-directed control in Experiment II -- Generalization (Subsection~\ref{sec:generalization}).
    \rev{Black areas represent obstacles.}}
    \label{fig:exp2_envs}
\end{figure}

We reuse the trained models from Experiment~\ref{sec:avoiding_obstacles}, and apply them for goal-directed control on two additional environments (see Figure~\ref{fig:exp2_envs}).
We only consider evolutionary-based active inference.

\begin{figure}[htb!]
    \centering
    \includegraphics[width=0.99\linewidth]{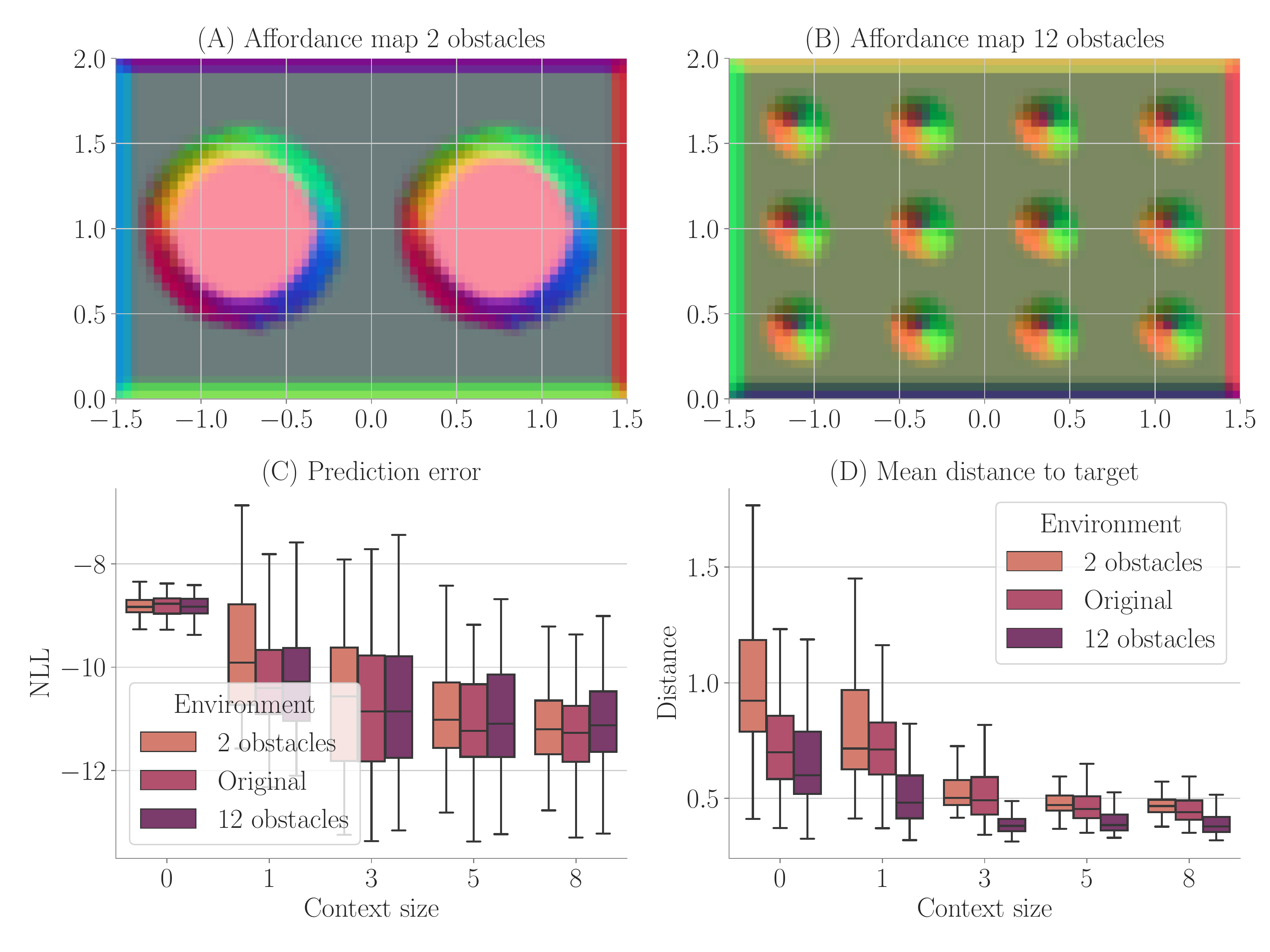}
    \caption{
        \rev{
        Results from Experiment II -- Generalization (Subsection~\ref{sec:generalization}).
        For each setting, the results are aggregated over $25$ differently initialized models which each performed $4$ goal-directed control runs for $200$ time steps.
        The box plots show the medians (horizontal bars), quartiles (boxes), and minima and maxima (whiskers).
        Data points outside of the range defined by extending the quartiles by $1.5$ times the interquartile range in both directions are ignored.
        "Original" refers to the environment from Experiment I (Subsection~\ref{sec:avoiding_obstacles}.
        (A) Exemplary affordance map for context size $8$ from environment with $2$ obstacles.
        To generate this map, we probed the environmental map at every sensible location, applied the vision model to each output, performed dimensionality reduction to $3$ via PCA, and interpreted the results as RGB values.
        (B) Exemplary affordance map for context size $8$ from environment with $12$ obstacles.
        (C) Prediction error during goal-directed control on all three environments.
        It is the negative log-likelihood of the actual change in position in the transition model's predicted probability distribution.
        (D) Mean distance to the target during goal-directed control on all three environments.
        }
    }
    \label{fig:exp2_results}
\end{figure}

Figure~\ref{fig:exp2_results} shows the results.
\rev{Prediction error and mean distance to target} (Figure~\ref{fig:exp2_results}C and D) indicate improvement \rev{with increasing context size}.
Furthermore, we find slightly worse performance in the environment with $2$ obstacles, while slightly better performance is achieved in the environment with $12$ obstacles.
We believe this is mainly due to the fact that the environment with \rev{$2$} obstacles blocks the direct path much more severely. 
Thus, overall these results indicate that (i) the system generalized well to similar environments and (ii) \rev{incorporating context codes} is beneficial for performance optimization.

\subsection{Experiment III: Behavioral-relevance of affordance codes} \label{sec:affordance_checks}

\begin{figure}[t!]
    \centering
    \includegraphics[width=0.5\linewidth]{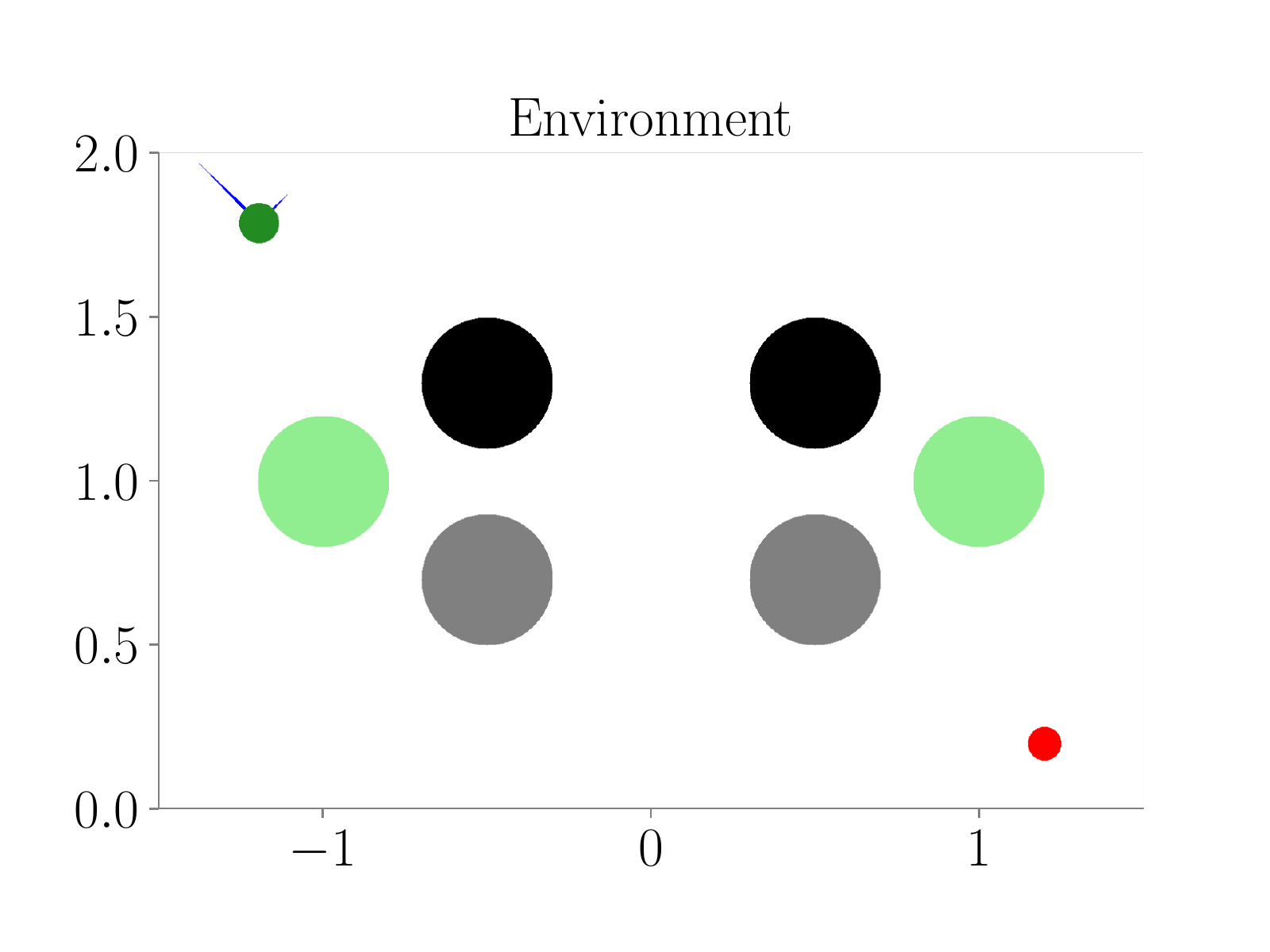}
    \caption{
        One of the environments (hard condition) used in Experiment III -- Behavioral-relevance of affordance codes (Subsection~\ref{sec:affordance_checks}).
        Black and grey circles represent obstacles.
        They look differently to the \rev{agent} but \rev{have} the same influence on behavior (i.e.\ path blockage). 
        Green circles \rev{are as well seen by the agent but} represent \rev{open area} and therefore mean the same a\rev{s} the white background behaviorally.
    }
    \label{fig:exp3_env}
\end{figure}

Affordances should only encode visual information if it is relevant to the behavior of an agent. 
Is our architecture able to ignore visual information for creating its affordance maps, if this information has no effect on the agent's behavior? 
Furthermore, affordances should encode different visual information with the same behavioral meaning similarly.
To investigate our architecture in this regard, we perform an experiment similar to Experiment I (Subsection~\ref{sec:avoiding_obstacles}), but with two additional channels in the cognitive map.
The first channel encodes the borders and upper obstacles, the second channel encodes the lower obstacles, and the third channel encodes meaningless information, which does not affect the behavior of the agent.
Figure~\ref{fig:exp3_env} shows the corresponding environment.
We compare the results from this ``hard condition'' to the results of Experiment I (Subsection~\ref{sec:avoiding_obstacles}), to which we refer as the ``easy condition''.
We only consider evolutionary-based active inference.

\begin{figure}[htb!]
    \centering
    \includegraphics[width=0.99\linewidth]{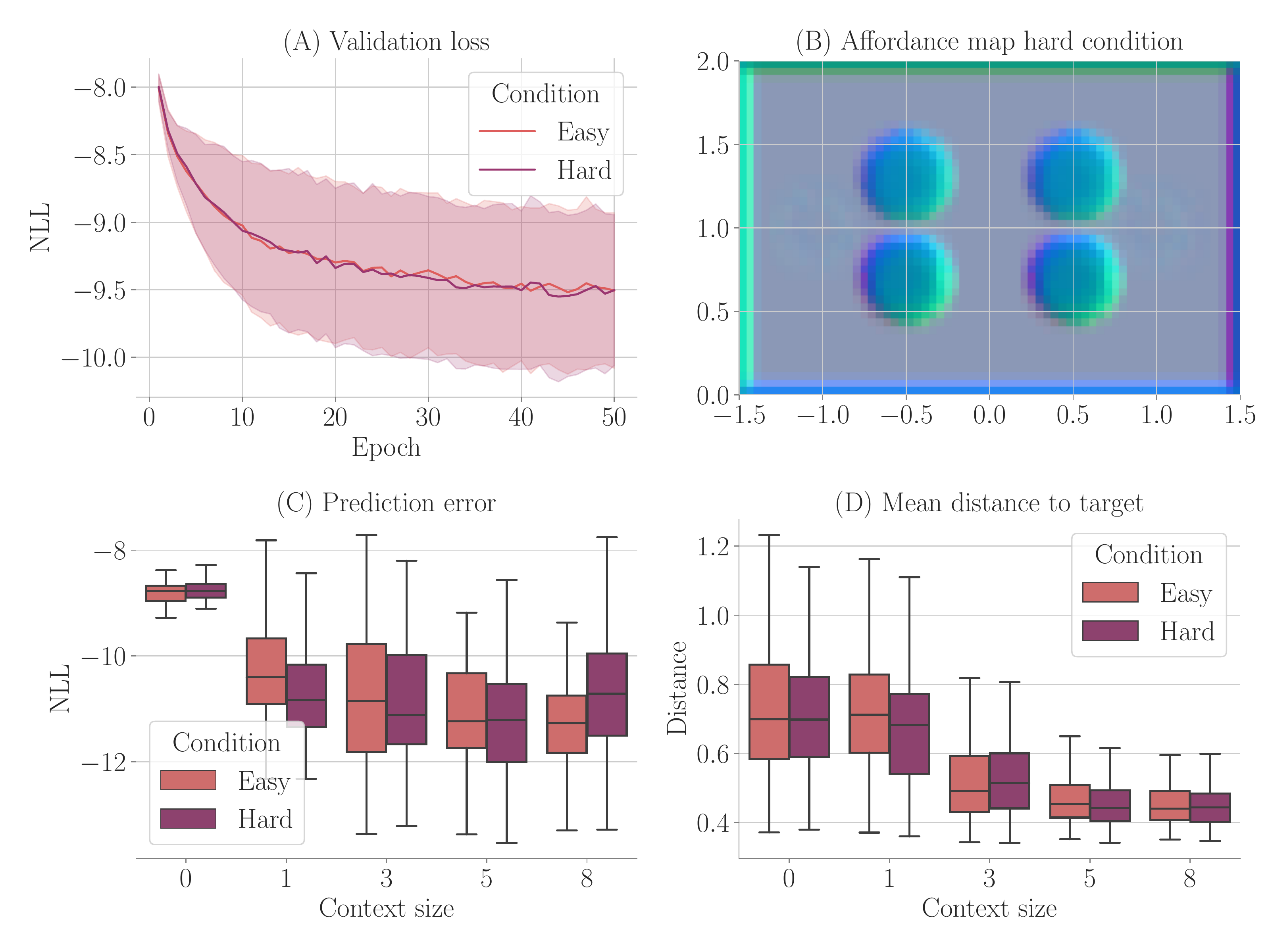}
    \caption{
        \rev{
        Results from Experiment III -- Behavioral-relevance of affordance codes (Subsection~\ref{sec:affordance_checks}).
        For each setting, the results are aggregated over $25$ differently initialized models which each performed $4$ goal-directed control runs for $200$ time steps.
        The box plots show the medians (horizontal bars), quartiles (boxes), and minima and maxima (whiskers).
        Data points outside of the range defined by extending the quartiles by $1.5$ times the interquartile range in both directions are ignored.
        "Easy" refers to the environment from Experiment I, "hard" refers to the condition with upper and lower obstacles encoded differently and additional meaningless information.
        (A) Validation loss during training.
        It is the negative log-likelihood of the actual change in position in the transition model's predicted probability distribution.
        Shaded areas represent standard deviations.
        (B) Exemplary affordance map for context size $8$ from environment with $2$ obstacles.
        To generate this map, we probed the environmental map at every sensible location, applied the vision model to each output, performed dimensionality reduction to $3$ via PCA, and interpreted the results as RGB values.
        (C) Prediction error during goal-directed control.
        It is the negative log-likelihood of the actual change in position in the transition model's predicted probability distribution.
        (D) Mean distance to the target during goal-directed control.
        }
    }
    \label{fig:exp3_results}
\end{figure}

Figure~\ref{fig:exp3_results} shows the results.
We do not find a significant difference between the two conditions.
The developing affordance map (Figure~\ref{fig:exp3_results}B) is qualitatively similar to the one obtained from Experiment~I (Subsection~\ref{sec:avoiding_obstacles}):
neither do significant visual differences between the encodings of the different obstacles remain, nor traces of the meaningless information.
Appendix~\ref{app:exp3} exemplarily shows how this affordance map develops over the course of training.
Finally, also performance \rev{in terms of prediction error and mean distance to the target} stays similar to Experiment~I when analyzing goal-directed control (Figure~\ref{fig:exp3_results}\rev{C and D}).

\subsection{Experiment IV: Uncertainty avoidance} \label{sec:avoiding_uncertainty}

Active inference \rev{considers uncertainty during} goal-directed control.
In this experiment, we examine the architecture's ability to avoid regions of uncertainty during planning.
We consider a run a success if the agent reached the target and was at no point inside a fog terrain.
As mentioned above, we introduce an additional hyperparameter $\beta$, which scales the influence of the entropy term on the free energy (see Equation~\ref{eq:efe}).
Here, we set $\beta$ to $10$ to foster avoidance of uncertainty.
We only consider evolutionary-based active inference.

We train the architecture on the environment depicted in Figure~\ref{fig:env}, this time black areas indicate fog terrains instead of obstacles.
The cognitive map consists of two channels:
one channel for fog terrains and one channel for the borders.

\begin{figure}[htb!]
    \centering
    \includegraphics[width=0.99\linewidth]{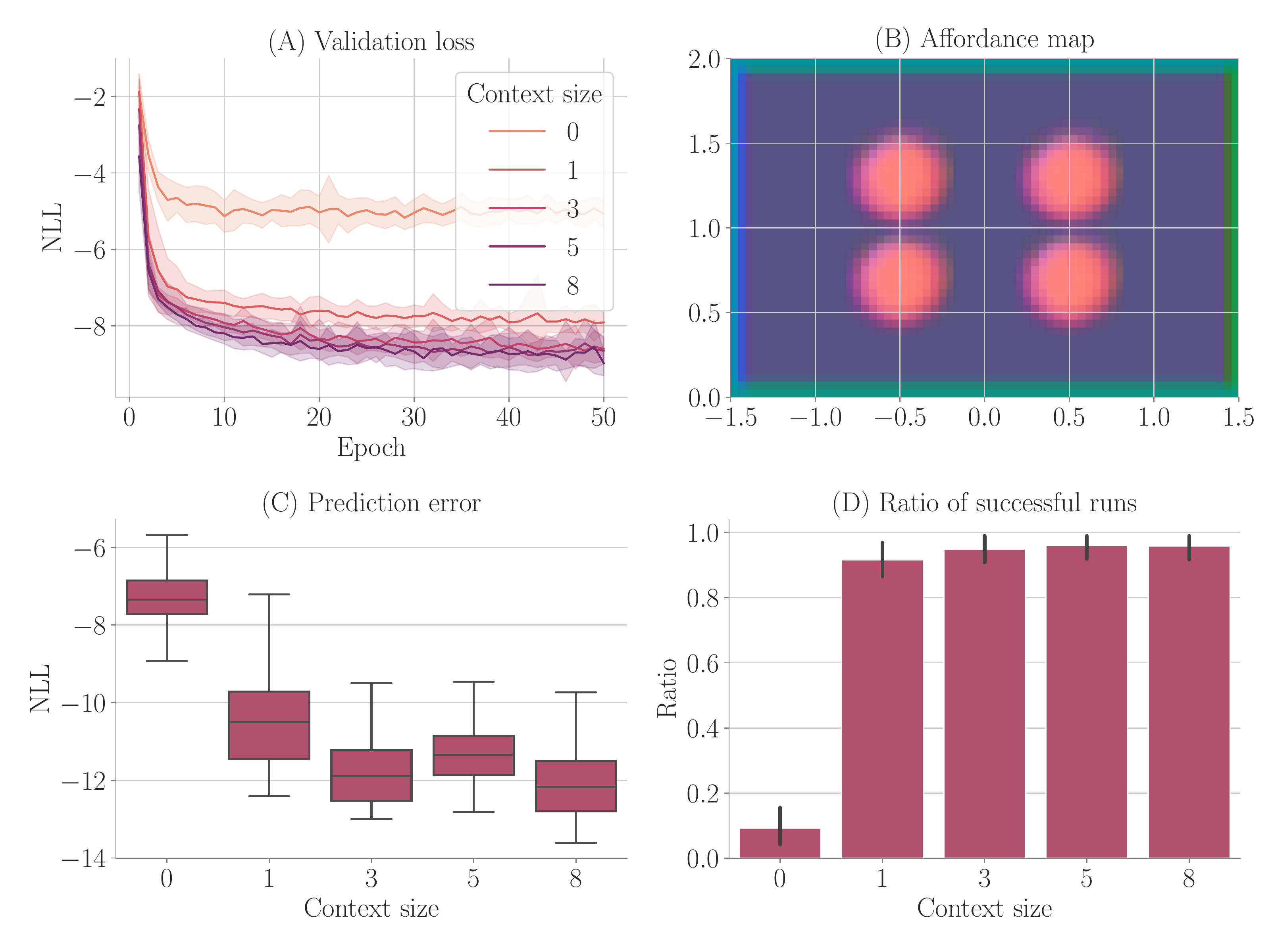}
    \caption{
        \rev{
        Results from Experiment IV -- Uncertainty avoidance (Subsection~\ref{sec:avoiding_uncertainty}).
        For each setting, the results are aggregated over $25$ differently initialized models which each performed $4$ goal-directed control runs for $200$ time steps.
        The box plot shows the medians (horizontal bars), quartiles (boxes), and minima and maxima (whiskers).
        Data points outside of the range defined by extending the quartiles by $1.5$ times the interquartile range in both directions are ignored.
        The bar plots show the means, where black lines represent the standard deviations.
        (A) Validation loss during training.
        It is the negative log-likelihood of the actual change in position in the transition model's predicted probability distribution.
        Shaded areas represent standard deviations.
        (B) Exemplary affordance map for context size $8$.
        To generate this map, we probed the environmental map at every sensible location, applied the vision model to each output, performed dimensionality reduction to $3$ via PCA, and interpreted the results as RGB values.
        (C) Prediction error during goal-directed control.
        It is the negative log-likelihood of the actual change in position in the transition model's predicted probability distribution.
        (D) Ratio of successful runs.
        A run was successful if the agent was closer to the target than $0.1$ units in at least one time step and did not touch fog in any time step.
        }
    }
    \label{fig:exp4_results}
\end{figure}

Figure~\ref{fig:exp4_results} shows the results.
We find that the context encoding clearly improves the validation loss (Figure~\ref{fig:exp4_results}A).
The affordance map (Figure~\ref{fig:exp4_results}B) shows that free areas, \rev{borders}, and fog are encoded differently.
\rev{The prediction error} (Figure~\ref{fig:exp4_results}C) improves when context is computed\rev{, while the ratio of successful runs (Figure~\ref{fig:exp4_results}D) stays relatively close to $1$}.

\subsection{Experiment V: Disentanglement} \label{sec:disentanglement}

In our final experiment, we examined the architecture's ability to combine previously learned affordance codes.
We trained each architecture instance on four different environments.
The environments are constructed as shown in Figure~\ref{fig:env}, black areas resembling obstacles in the first, fog terrains in the second, force fields pointing upwards in the third, and force fields pointing downwards in the fourth environment.
Accordingly, the cognitive map consists of four channels---one channel for each of the aforementioned properties.
We evaluate the architecture on procedurally generated environments.
In each environment, a randomly chosen amount between $6$ and $10$ obstacles, fog terrains, force fields pointing downwards, and force fields pointing upwards with randomly chosen radii between $0.1$ and $0.5$ are placed at random locations in the environment.
All obstacles and fog terrains have a minimum distance of $0.15$ units from each other, ensuring that the agent is able to fly between them---thus prohibiting dead ends.
Furthermore, all properties have a minimum distance of $0.15$ to each border, again to avoid dead ends.
Patches of size $0.4 \times 0.4$ units are left free in the corners such that start and target positions are not affected.
We generate environments with two different conditions.
In the first condition (easy), force fields are handled similarly to obstacles and fog terrains in the way that they have a minimum distance of $0.15$ units to all other obstacles, terrains, and force fields.
This means that properties do not overlap.
In the second condition (hard), force fields can overlap with each other, obstacles, and fog terrains.
We only consider evolutionary-based active inference.
\rev{In addition to the context sizes from before, we also evaluate the architecture for context sizes $16$ and $32$.}

\begin{figure}[htb!]
    \centering
    \includegraphics[width=.99\linewidth]{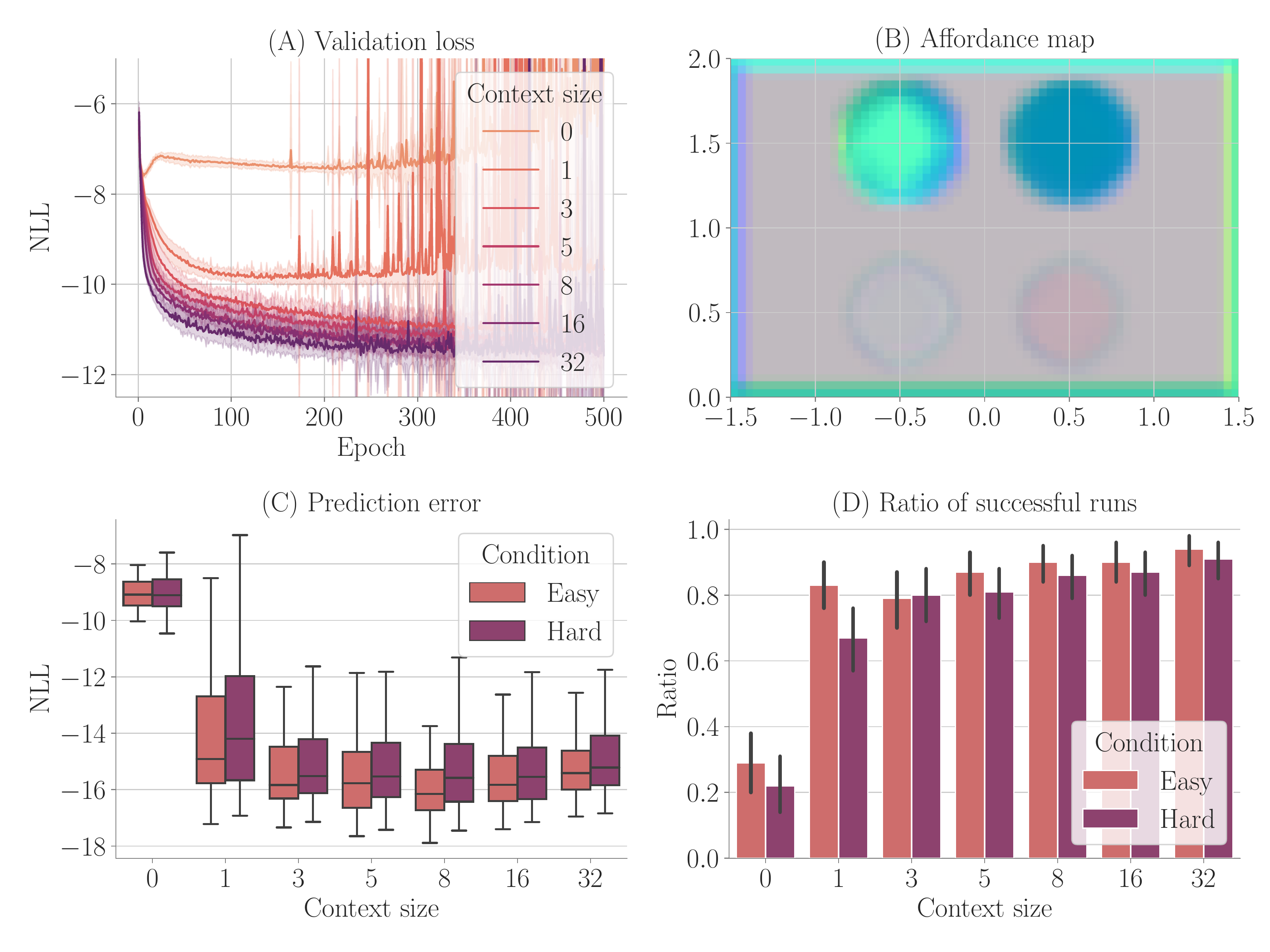}
    \caption{
        \rev{
        Results from Experiment V -- Disentanglement (Subsection~\ref{sec:disentanglement}).
        For each setting, the results are aggregated over $25$ differently initialized models which each performed $4$ goal-directed control runs for $200$ time steps.
        The box plot shows the medians (horizontal bars), quartiles (boxes), and minima and maxima (whiskers).
        Data points outside of the range defined by extending the quartiles by $1.5$ times the interquartile range in both directions are ignored.
        The bar plot shows the means, where black lines represent the standard deviations.
        "Easy" refers to the condition where obstacles and terrains do not overlap, "Hard" refers to the condition where, force fields can overlap with each other, obstacles, and fog terrains.
        (A) Validation loss during training.
        It is the negative log-likelihood of the actual change in position in the transition model's predicted probability distribution.
        Shaded areas represent standard deviations.
        (B) Exemplary affordance map for context size $8$.
        This environment contains all four proporties (obstacle in the upper left, for terrains in the upper right, force field pointing downwards in the lower left, and force field poiting upwards in the lower right).
        It was not used during goal-directed control.
        To generate this map, we probed the environmental map at every sensible location, applied the vision model to each output, performed dimensionality reduction to $3$ via PCA, and interpreted the results as RGB values.
        (C) Prediction error during goal-directed control.
        It is the negative log-likelihood of the actual change in position in the transition model's predicted probability distribution.
        (D) Ratio of successful runs.
        A run was successful if the agent was closer to the target than $0.1$ units in at least one time step and did not touch fog in any time step.
        }
    }
    \label{fig:exp5_results}
\end{figure}

Figure~\ref{fig:exp5_results} shows the results.
Larger context sizes lead to smaller validation losses (Figure~\ref{fig:exp5_results}A).
\rev{The affordance map computed on an environment containing all properties (Figure~\ref{fig:exp5_results}B) shows that the network has learned to encode the distinct areas indeed with distinct encodings.}
The encoding also incorporates boundary directions, thus encoding the properties relative to the free space from which the agent may enter the area (see, for example, the borders of the environment).
As expected, the agent always performs better in the easy condition \rev{(Figure~\ref{fig:exp5_results}C and D)}.
\rev{In both conditions, performance improves with increasing context size.}

\section{Discussion}

In this paper, we have connected active inference with the theory of affordances in order to guide the search for suitable behavioral policies via active inference in recurrent neural networks.
The resulting architecture is able to perform goal-directed planning while considering the properties of the agent's local environment.
This chapter provides a summary of our architecture's abilities, compares it to related work, and eventually presents possible future work directions.

\subsection{Conclusion}

Experiment I (Subsection~\ref{sec:avoiding_obstacles}) \rev{showed} that our proposed architecture facilitates goal-directed planning via active inference.
Both the validation loss as well as performance during goal-directed control revealed an advantage of incorporating affordance information, i.e.\ using a context size larger than $0$.
The affordance maps confirmed that the architecture is able to infer relationships between environmental features and their meaning for the agent's behavior:
Depending on the direction of and the distance to the next obstacle, different codes emerged.
\rev{We assume that context size larger than $1$ allows an easier encoding of the direction of and distance to the obstacles in relation to the agent.}
\rev{Evolutionary-based active inference outperforms gradient-based active if a context is used.
This is due to the fact that gradient-based active inference relies on the gradients being backpropagated through the predicted sequence of sensory states.
These gradients cannot be backpropagated through the context codes, since these depend on the visual information which the model acquires via a look-up in the environmental map.
Therefore, vital information is missing during gradient-based active inference in order to optimize the policy accordingly.}
Experiment II (Subsection~\ref{sec:generalization}) \rev{showed} that once the relationship between environmental features and their meaning was learned, this knowledge can be generalized to other environments with similar, but differently sized and positioned obstacles \rev{of different amounts}. 
Experiment III (Subsection~\ref{sec:affordance_checks}) \rev{showed} that our architecture is able to map properties of the environment that are encoded differently visually but have the same influence on behavior onto the same affordance codes.
\rev{This matches our general definition of affordances, namely an affordance encoding locally behavior-modifying properties of the environment.
In the future, we plan to evaluate our architecture in environments where the connection to task-relevancy is more concrete.
An example would be a key that needs to be picked up by the agent in order for a door to be encoded as passable.}
Experiment IV (Subsection~\ref{sec:avoiding_uncertainty}) \rev{showed} that our architecture is able to avoid regions of uncertainty (fog terrains) during planning via active inference.
Experiment V (Subsection~\ref{sec:disentanglement}) \rev{emphasized} our architecture's generalization abilities, but also showed that the learned affordances \rev{are not disentangled}.
If properties of the environment do not overlap and with sufficiently large context size, the agent can successfully reach the target without touching regions of uncertainty nearly all the time.
This is less so \rev{if} properties do overlap.
We propose that an additional regularization that may foster a disentanglement or factorization of the learned affordances could lead to a fully successful generalization to arbitrary combinations of previously encountered properties.

\revtwo{
Our notion of affordances admittedly slightly differs from the original definition in \citet{gibson1986ecological}.
In this work, we assume that every action is possible everywhere, but that only the effects differ depending on the environmental context.
This was certainly the case in the environment we used in our experiments.
We think that this is also often the case in the real world---particularly when actions are considered on the lowest level only, i.e. muscle movements.
When increasing behavioral abstraction, though, this might not necessarily be the case any longer.
For example, the high-level, composite action of driving a nail into a wall clearly is not possible in every environmental context.
In the future, we want to investigate how our architecture can be expanded to flexibly and hierarchically process event-like structures \citep{EventPerceptioZacks2007,EventPredictiveButz2021,zacks2001event,Eppe:2022}.
In order to foster these event structures, inductive biases as in \cite{butz_learning_2019} and \cite{Gumbsch:2021b} might be necessary, which assume that most of the time agents are within ongoing events and that event boundaries characterize transitions between events. 
Such event models may thus set the general context. 
The proposed vision model may then be conditioned on this context to enable accurate, event-conditioned action-effect predictions. 
As a result, the event-conditioned model would learn to encode when the action result that is associated with a particular event-specific affordance can be accomplished.
}

\revtwo{
We treat the context size in our experiments as a hyperparameter, which needs to be set by the experimenter.
Our results show that when the context size is too small, the system is unable to learn all environmental influences on action effects.
Larger context sizes, on the other hand, tend to decrease prediction error.
Improvement does not only depend on the context size, though, but also on the computational capacity of the vision and transition models.
What context size is necessary for a certain number of possible environmental interactions remains an open question for future research.
One possible direction here is to let the model adapt the context size on demand.
In this case, the computational capacity of the vision and transition models need to be adapted as well, which poses a challenge.
Furthermore, the transition model may infer, if equipped with recurrences to deal with partial observability, information that would otherwise be immediately available via the vision module.
This leads to a competition during learning, which needs to be studied further.
}

\rev{Even though our architecture was successful in solving the presented tasks, it clearly has some limitations.
First, our model computes affordance codes directly from visual information.
Since it is not able to memorize which affordances are where, it constantly needs to perform look-ups in the environmental map.
Second, our proposed architecture solves the considered tasks in a greedy manner.
During planning, we compare the predicted sensory states to a fixed desired sensory state over the predicted trajectory.
Our model therefore prefers actions that lead closer to the target only within the prediction horizon.
This can be problematic if we consider e.g. tool use.
Imagine an environment with keys and doors.
Here, it might be necessary to temporarily steer away from the target in order to pick up a key and eventually get closer to the target after unlocking and passing through the door.
Without further modifications, our agent would not make such a detour deliberately.
In the future work section below, we make suggestions on how these limitations may be overcome.}

\rev{Reinforcement learning (RL) \citep{sutton2018reinforcement} is another popular approach for solving POMDPs.
Therefore, in future work, it would be interesting to see how an RL agent performs in comparison to our agent.
A central aspect of our architecture is the look-up in the environmental which makes the emergence of affordance maps possible.
While certainly possible, it is not straight-forward how this would be implemented in a classical RL agent.
Classical RL agents do not predict positional changes which are necessary for the look-up.
Furthermore, it was shown that RL agents struggle with offline learning \citep{levine2020offline} and generalization to similar environments \citep{cobbe2019quantifying}.}

\subsection{Future work}

\rev{In this work, we trained our architecture on previously collected data and only afterwards performed goal-directed control.
Alternatively, one could perform goal-directed control from the very beginning and train the architecture on inferred actions and the corresponding encountered observations in a self-supervised learning manner.
This should increase performance since the distribution of the training data for the transition model then more closely matches the distribution of the data encountered during control.
In this case, the exploration-exploitation dilemma needs to be resolved:
How should the agent decide whether it should exploit previously acquired knowledge to reach its goal or instead explore the environment to gain further knowledge that can be exploited in later trials?
The active inference mechanism 
\citep{Friston:2015} generally provides a solution to this problem, although optimal parameter tuning remains challenging \citep{Tani:2017a}.}

Future work could \rev{also} examine to what extent it is possible to fully memorize affordance\rev{s} akin to a cognitive map.
A straightforward approach would be to train a multi-layer perceptron that maps absolute positions onto affordance codes, in which case translational invariance is lost.
Alternatively, a recurrent neural network that receives actions could predict affordances in future time steps conditioned upon previously encountered affordances.
Additionally, the \rev{transition} model could be split into an encoder, which maps sensory states onto internal hidden states, and a transition model, which maps internal hidden states and actions onto next internal hidden states.
The introduction of an observation model that translates internal hidden states back into sensory states would then enable the whole planning process to take place in hidden state space akin to PlaNet~\citep{PlaNet} and Dreamer~\citep{1912.01603v3}.

\section*{Funding}
This research was supported by the German Research Foundation (DFG) within Priority-Program SPP 2134 - project “Development of the agentive self” (BU 1335/11-1, EL 253/8-1), the research unit 2718 on “Modal and Amodal Cognition: Functions and Interactions´´ (BU 1335/12-1), and the Machine Learning Cluster of Excellence funded by the Deutsche Forschungsgemeinschaft (DFG, German Research Foundation) under Germany’s Excellence Strategy – EXC number 2064/1 – Project number 390727645. Finally, additional support came
from the Open Access Publishing Fund of the University of Tübingen.

\rev{\section*{Acknowledgements}
The authors thank the International Max Planck Research School for Intelligent Systems (IMPRS-IS) for supporting Fedor Scholz and Christian Gumbsch.}

\bibliographystyle{unsrtnat} 
\bibliography{AffordancesAndActiveInference_arxiv.bib}


\appendix

\section*{Appendix}

\section{Model and training hyperparameters and details}\label{app:model}

\rev{Unless noted otherwise, we use the hyperparameters specified in this section.}
Visual input \rev{$\mathbf{v}$} consists of $11 \times 11$ pixels with the number of channels depending on the experiment.
We obtain it by rasterization of a $0.5 \times 0.5$ square of the environment, centering and excluding the agent.
Due to the maximum velocity of $0.23$ units per time step, the agent's next position is always within it's visual field.
Each channel corresponds to a property (obstacle, fog terrain, force field up and down) of the environment.
The presence of a property is encoded with $1$s, while the rest of the tensor is set to $0$.

\rev{The vision model $v_M$ is given by a CNN.}
\rev{It} consists of a convolutional layer, a max pooling layer, and another convolutional layer followed by a fully connected layer.
The convolutional layers have kernel size $3 \times 3$ with stride $1$, no padding, $4$ channels \rev{if dim$(\mathbf{c}) < 32$, and $8$ channels if dim$(\mathbf{c}) = 32$}.
The max pooling layer has a receptive field size of $3 \times 3$ with stride $2$.
The fully connected layer has size $8$ \rev{if dim$(\mathbf{c}) < 5$, $16$ if $5 < \text{dim}(\mathbf{c}) \leq 16$, and $32$ if dim$(\mathbf{c}) = 32$}.
We use the $\tanh$ activation function in all layers.
The vision model has
\rev{$$564 + \text{dim}(\mathbf{i}) \cdot 36 + \text{dim}(\mathbf{c}) \cdot 9 \text{, if dim}(\mathbf{c}) < 5$$
$$964 + \text{dim}(\mathbf{i}) \cdot 36 + \text{dim}(\mathbf{c}) \cdot 17 \text{, if } 5 < \text{dim}(\mathbf{c}) \leq 16$$
$$4312 + \text{dim}(\mathbf{i}) \cdot 72 \text{, if dim}(\mathbf{c}) = 32$$}
parameters in total, where $\text{dim}(\mathbf{i})$ denotes the number of channels of the input.
We use Adam \citep{kingma_adam_2014} as our optimizer with learning rate $0.\rev{0}0075$, $\beta$-values $(0.9, 0.999)$, and $\epsilon = 1e-4$.
\rev{In Experiment V, we use a $10$th of the learning rate.}
We perform gradient clipping \citep{GradientNormClipping} and set the maximum norm to $2$.

\rev{The transition model $t_M$ is given by a MLP.
The first fully connected layer has hidden size $32$ with biases turned off.
It is followed by the $\tanh$ activation function.
The first of the two parallel fully connected layers predicts mean vectors with the linear activation function.
A second parallel fully connected layer predicts vectors of standard deviations via the exponential activation function, providing non-negative values and therefore ensuring valid standard deviations.
From a probabilistic point of view, under the assumption that the values before the activation function are uniformly distributed, these mappings implement an uninformative prior in a Bayesian framework \citep{SimplifyingNeu}.
The changes in position are scaled up by a constant factor of $4$ before feeding them into $t_M$, such that it receives inputs which approximately cover the interval between $-1$ and $1$.
The transition model has
$$324 + \text{dim}(\mathbf{c}) \cdot 32$$
parameters in total.
We use Adam \citep{kingma_adam_2014} as our optimizer with learning rate $0.008$, $\beta$-values $(0.9, 0.999)$, and $\epsilon = 1e-4$.
In Experiment V, we use a $10$th of the learning rate.
We perform gradient clipping \citep{GradientNormClipping} and set the maximum norm to $1.2$.}

We generate training data by sending randomly generated actions to the environment.
Actions were generated in a way that ensures good coverage of the whole environment.
For each environment used in our experiments, we gather $200$ sequences of sensor-action-tuples.
We use $160$ sequences for training and $40$ for validation.
Each sequence has a length of $300$ time steps.
\rev{We train both components jointly end-to-end with batch size $10$ for $50$ epochs in Experiments I-IV and for $500$ epochs in Experiment V.}
We backpropagate the error through time every $50$ time steps and reset the hidden states every $7$ batches.
This way we train the model to avoid exploding hidden states also during goal-directed control.

\section{Details on planning algorithms}
\label{app:planning}

When planning with gradient-based active inference, we apply the following adjustments to improve performance.
Firstly, if an optimization cycle increases EFE, we perform early stopping and use the policy from the cycle before.
Secondly, we decrease the learning rate exponentially over the policy from the future to the present.
This leads to more stable paths since actions which lie in the later future are adapted more than actions to be executed in the nearer future.
More precisely, given a mean learning rate $\alpha$ and decay $\gamma$, we set the learning rate for action $\mathbf{a}^{t+\tau}$ to:
$$\alpha_{\mathbf{a}^{t+\tau}} = \alpha \cdot \frac{\gamma^{P-\tau}}{\sum_\tau^P \gamma^{P-\tau}}$$
See Appendix~\ref{app:federivative} for a description of how to compute gradients when the objective is given by the FE between two multivariate normal distributions.
After each update, we clamp the policy to be in the correct value range.
Finally, after optimization, we shift the policy while copying the last element.
We use \rev{stochastic gradient descent} with learning rate $0.005$, set the exponential learning rate decay to $\gamma= 0.9$, and perform $\rev{5}0$ optimization cycles.
\rev{If a policy update leads to worse performance, we stop the optimization and use the policy from before.}

During evolutionary-based planning, we use normal distributions to model actions.
In order to improve performance, we apply the following modifications.
We use a momentum term on the means and covariances \citep{de2005tutorial}.
After a single optimization iteration, we keep a fixed number $K$ of the elites for the next iteration \citep{iCEM}.
After optimization, we do not discard the means but shift them \citep{wang2019exploring, PETS} while copying the last action in order to not start from scratch in the next optimization.
We reset the variances, however, to avoid local minima.
Analogously, we shift the elites that we keep \citep{iCEM}.
We use the first action from the best sampled policy as the optimization result \citep{iCEM}.
Instead of clipping sampled actions, we perform rejection sampling and sample until we have an action within the allowed value range.
We generate $50$ trajectory candidates, use $5$ elites for parameters estimation, keep $K=2$ elites for the next optimization cycle, use an initial covariance of $0.5$, and a momentum of $0.1$.

\section{Affordance maps from experiment III after different amounts of epochs}
\label{app:exp3}

\begin{figure}[htb!]
    \centering
    \includegraphics[width=0.99\linewidth]{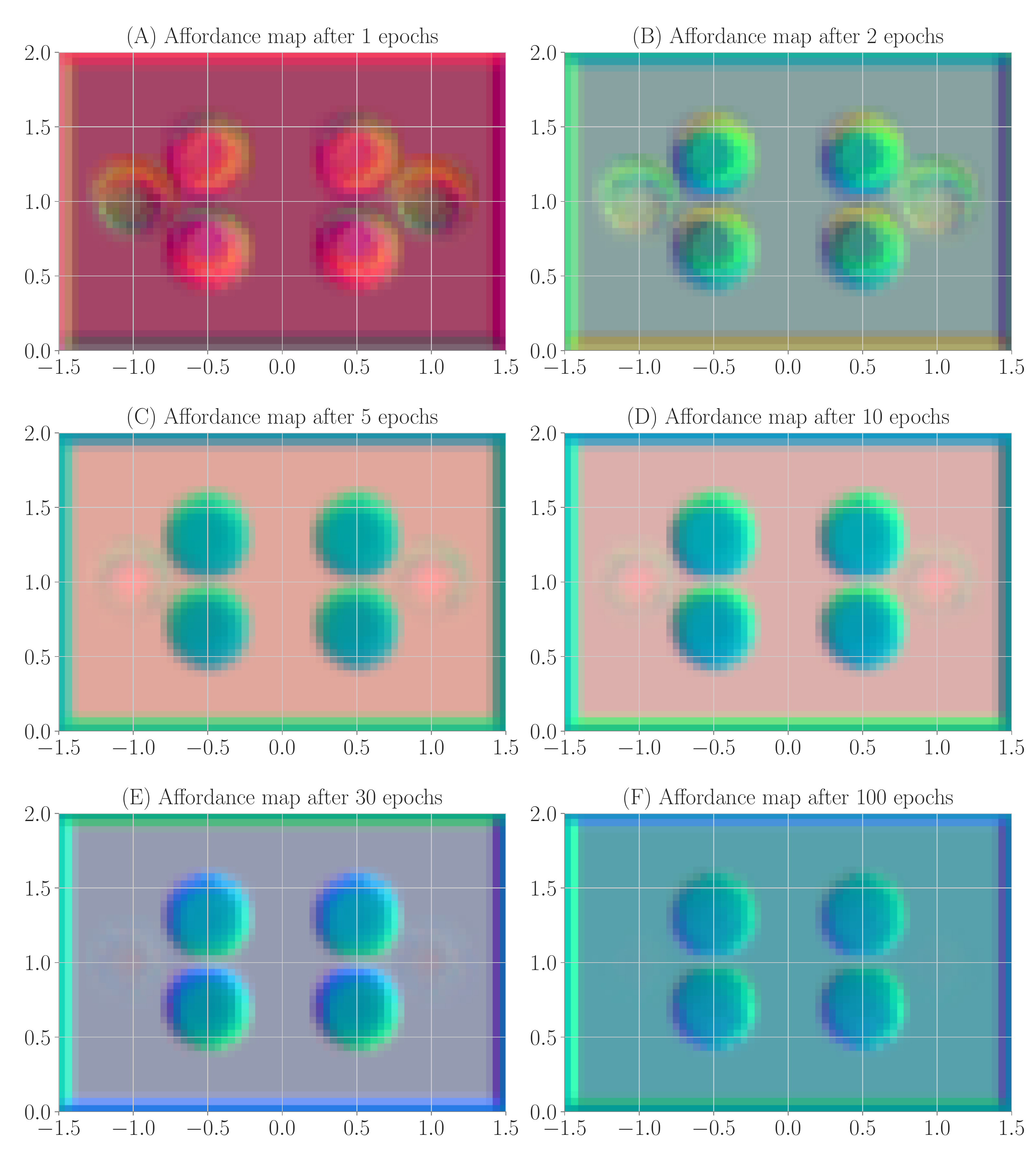}
    \caption{
        Exemplary affordance maps from Experiment III (Subsection~\ref{sec:affordance_checks}) \rev{for context size $8$} after different amounts of epochs.
        \rev{To generate these maps, we probed the environmental map at every sensible location, applied the vision model to each output, performed dimensionality reduction to $3$ via PCA, and interpreted the results as RGB values.}
    }
    \label{fig:exp3_results_maps}
\end{figure}

Here, we show affordance maps from Experiment III (Subsection~\ref{sec:affordance_checks}) after different amounts of epochs.
In Figure~\ref{fig:exp3_results_maps} we see that with increasing amounts of training epochs, the upper and lower obstacles get encoded more similarly, the additional meaningless information gets more filtered out, and the affordance maps get more distinctive regarding the encoding of different behavioral possibilities.

%
%

\section{Derivative of negative log-likelihood in a normal distribution} \label{app:nllderivative}

In this section we derive the negative log-likelihood in a multivariate normal distribution with respect to the distribution's parameters.

The likelihood in a multivariate normal distribution is given by its probability density function:
\begin{equation}
    \begin{split}
        \mathcal{L} & = p(\mathbf{x} \mid \mathbf{\mu}, \mathbf{\Sigma}) \\
                    & = \mathcal{N}(\mathbf{x} \mid \mathbf{\mu}, \mathbf{\Sigma}) \\
                    & = \frac{1}{\sqrt{(2\pi)^k |\mathbf{\Sigma}|}} \cdot e^{-\frac{1}{2} (\mathbf{x} - \mathbf{\mu})^T \mathbf{\Sigma}^{-1} (\mathbf{x} - \mathbf{\mu})}
    \end{split}
\end{equation}

This leads to the following log-likelihood:
\begin{equation}
    \begin{split}
        \mathcal{LL} & = \log 1 - \log \sqrt{(2 \pi)^k |\mathbf{\Sigma}|} -\frac{1}{2} (\mathbf{x} - \mathbf{\mu})^T \mathbf{\Sigma}^{-1} (\mathbf{x} - \mathbf{\mu}) \\
                     & = - \frac{1}{2} (k \cdot \log (2 \pi) + \log |\mathbf{\Sigma}| + (\mathbf{x} - \mathbf{\mu})^T \mathbf{\Sigma}^{-1} (\mathbf{x} - \mathbf{\mu})) \\
    \end{split}
\end{equation}

Now, we take the derivative of the log-likelihood function with respect to the parameters of our probability distribution.
The resulting quantity is also referred to as the \emph{score}.
We start by calculating the derivative with respect to the mean $\mathbf{\mu}$:
\begin{equation}
    \begin{split}
        \frac{\partial \mathcal{LL}}{\partial \mathbf{\mu}} & = \frac{\partial}{\partial \mathbf{\mu}} - \frac{1}{2} (k \cdot \log (2 \pi) + \log |\mathbf{\Sigma}| + (\mathbf{x} - \mathbf{\mu})^T \mathbf{\Sigma}^{-1} (\mathbf{x} - \mathbf{\mu})) \\
                                                   & = - \frac{1}{2} (\frac{\partial (\mathbf{x} - \mathbf{\mu})^T \mathbf{\Sigma}^{-1} (\mathbf{x} - \mathbf{\mu})}{\partial \mathbf{\mu}} ) \\
                                                   & = \mathbf{\Sigma}^{-1}(\mathbf{x} - \mathbf{\mu}) \\
    \end{split}
\end{equation}

Next, we calculate the derivative with respect to the covariance matrix $\mathbf{\Sigma}$.
We assume $\mathbf{\Sigma}$ to be symmetric:
\begin{equation}
    \begin{split}
        \frac{\partial \mathcal{LL}}{\partial \mathbf{\Sigma}} & = \frac{\partial}{\partial \mathbf{\Sigma}} - \frac{1}{2} (k \cdot \log (2 \pi) + \log |\mathbf{\Sigma}| + (\mathbf{x} - \mathbf{\mu})^T \mathbf{\Sigma}^{-1} (\mathbf{x} - \mathbf{\mu})) \\
                                                      & = - \frac{1}{2} (\frac{\partial \log |\mathbf{\Sigma}|}{\partial \mathbf{\Sigma}} + \frac{\partial (\mathbf{x} - \mathbf{\mu})^T \mathbf{\Sigma}^{-1} (\mathbf{x} - \mathbf{\mu})}{\partial \mathbf{\Sigma}}) \\
                                                      & = - \frac{1}{2} (\mathbf{\Sigma}^{-1} - \mathbf{\Sigma}^{-1} (\mathbf{x} - \mathbf{\mu})(\mathbf{x} - \mathbf{\mu})^T \mathbf{\Sigma}^{-1}) \\
    \end{split}
\end{equation}

We are now able to calculate the derivatives of the log-likehood function of a multivariate normal distribution.
In this work, we applied two simplifications:
First, we used the special case of a bivariate normal distribution.
Second, we assume all covariances to be $0$, leading to a diagonal covariance matrix.
With these assumptions, the multivariate normal distribution factors into two univariate normal distributions.
We replace the covariance matrix $\mathbf{\Sigma}$ with a vector of variances $\mathbf{\sigma}^2$ and end up with the following score:
\begin{equation}
    \begin{split}
        \frac{\partial \mathcal{LL}}{\partial \mu_i} & = \frac{x_i - \mu_i}{\sigma_i^2} \\
        \frac{\partial \mathcal{LL}}{\partial \sigma^2_i} & = \frac{1}{2\sigma_i^2} (\frac{1}{\sigma_i^2} (x_i - \mu_i)^2 - 1) \\
    \end{split}
\end{equation}

\section{Derivative of free energy between normal distributions} \label{app:federivative}

In this section we derive the expected free energy as used in this work between two multivariate normal distributions with respect to the parameters of one of the distributions.
We first take the derivative of the entropy and subsequently of the divergence term.

The entropy of a multivariate normal distribution is given by:
\begin{equation}
    \begin{split}
        H[p(\mathbf{x}|\mathbf{\mu}, \mathbf{\Sigma})] & = \frac{1}{2} \log |2 \pi e \mathbf{\Sigma}| \\
            & = \frac{1}{2} (\log (2 \pi e)^k + \log |\mathbf{\Sigma}|) \\
    \end{split}
\end{equation}

Now we take the derivative with respect to the mean $\mathbf{\mu}$:
\begin{equation}
    \begin{split}
        \frac{\partial}{\partial \mathbf{\mu}} H[p(\mathbf{x} \mid \mathbf{\mu}, \mathbf{\Sigma})] & = \frac{\partial}{\partial \mathbf{\mu}} \frac{1}{2} (\log (2 \pi e)^k + \log |\mathbf{\Sigma}|) \\
        & = 0 \\
    \end{split}
\end{equation}

Next, we take the derivative with respect to the covariance matrix $\mathbf{\Sigma}$ (assuming $\mathbf{\Sigma}$ to be symmetric):
\begin{equation}
    \begin{split}
        \frac{\partial}{\partial \mathbf{\Sigma}} H[p(\mathbf{x} \mid \mathbf{\mu}, \mathbf{\Sigma})] & = \frac{\partial}{\partial \mathbf{\Sigma}} \frac{1}{2} (\log (2 \pi e)^k + \log |\mathbf{\Sigma}|) \\
        & = \frac{1}{2} \frac{\partial \log |\mathbf{\Sigma}|}{\partial \mathbf{\Sigma}} \\
        & = \frac{1}{2} \mathbf{\Sigma}^{-1} \\
    \end{split}
\end{equation}

We are now able to calculate the derivative of the entropy of a multivariate normal distribution.
Following the simplifications from above (Section~\ref{app:nllderivative}), we again replace the covariance matrix $\mathbf{\Sigma}$ with a vector of variances $\mathbf{\sigma}^2$ and end up with the following gradients:
\begin{equation}
    \begin{split}
        \frac{\partial H[p(\mathbf{x} \mid \mathbf{\mu}, \mathbf{\sigma})]}{\partial \mu_i} & = 0 \\
        \frac{\partial H[p(\mathbf{x} \mid \mathbf{\mu}, \mathbf{\sigma})]}{\partial \sigma_i^2} & = \frac{1}{2} \sigma_i^{-2}
    \end{split}
\end{equation}

The Kullback-Leibler divergence between to multivariate normal distributions is given by:
\begin{equation}
    \begin{split}
        D[p(\mathbf{x}_0 \mid \mathbf{\mu}_0, \mathbf{\Sigma}_0)  \mid \mid  p(\mathbf{x}_1 \mid \mathbf{\mu}_1, \mathbf{\Sigma}_1)] = &\frac{1}{2} (\text{tr} (\mathbf{\Sigma}_1^{-1} \mathbf{\Sigma}_0) \\
        & + (\mathbf{\mu}_1 - \mathbf{\mu}_0)^T \mathbf{\Sigma}_1^{-1} (\mathbf{\mu}_1 - \mathbf{\mu}_0) \\
        & - k - \log \frac{|\mathbf{\Sigma}_1|}{|\mathbf{\Sigma}_0|} )
    \end{split}
\end{equation}

We first take the derivative with respect to the mean of the first distribution $\mathbf{\mu}_0$:
\begin{equation}
    \begin{split}
        \frac{\partial}{\partial \mathbf{\mu}_0} D[p(\mathbf{x}_0 \mid \mathbf{\mu}_0, \mathbf{\Sigma}_0) \mid \mid  p(\mathbf{x}_1 \mid \mathbf{\mu}_1, \mathbf{\Sigma}_1)] & = \frac{\partial}{\partial \mathbf{\mu}_0} \frac{1}{2} (\mathbf{\mu}_1 - \mathbf{\mu}_0)^T \mathbf{\Sigma}_1^{-1} (\mathbf{\mu}_1 - \mathbf{\mu}_0) \\
        & = - \mathbf{\Sigma}_1^{-1} (\mathbf{\mu}_1 - \mathbf{\mu}_0) \\
    \end{split}
\end{equation}

Now we take the derivative with respect to the covariance matrix of the first distribution $\mathbf{\Sigma}_0$ (assuming $\mathbf{\Sigma}_0$ and $\mathbf{\Sigma}_1$ to be symmetric):
\begin{equation}
    \begin{split}
        \frac{\partial}{\partial \mathbf{\Sigma}_0} D[p(\mathbf{x}_0 \mid \mathbf{\mu}_0, \mathbf{\Sigma}_0) \mid \mid  p(\mathbf{x}_1 \mid \mathbf{\mu}_1, \mathbf{\Sigma}_1)] & = \frac{\partial}{\partial \mathbf{\Sigma}_0} \frac{1}{2} (\text{tr} (\mathbf{\Sigma}_1^{-1} \mathbf{\Sigma}_0) + \log \frac{|\mathbf{\Sigma}_1|}{|\mathbf{\Sigma}_0|}) \\
        & = \frac{1}{2} (\mathbf{\Sigma}_1^{-1} - \mathbf{\Sigma}_0^{-1}) \\
    \end{split}
\end{equation}

We are now able to calculate the gradients of the Kullback-Leibler divergence between two multivariate normal distributions.
Following the simplifications from above, we again replace the covariance matrices $\mathbf{\Sigma}_0$ and $\mathbf{\Sigma}_1$ with vectors of variances $\mathbf{\sigma}_0^2$ and $\mathbf{\sigma}_1^2$ and end up with the following gradients:
\begin{equation}
    \begin{split}
        \frac{\partial D[p(\mathbf{x}_0 \mid \mathbf{\mu}_0, \mathbf{\Sigma}_0) \mid \mid p(\mathbf{x}_1 \mid \mathbf{\mu}_1, \mathbf{\Sigma}_1)]}{\partial \mu_{0,i}} & = -\sigma_{1,i}^{-2} (\mu_{1,i} - \mu_{0,i}) \\ 
        \frac{\partial D[p(\mathbf{x}_0 \mid \mathbf{\mu}_0, \mathbf{\Sigma}_0) \mid \mid p(\mathbf{x}_1 \mid \mathbf{\mu}_1, \mathbf{\Sigma}_1)]}{\partial \sigma_{0,i}^2} & = \frac{1}{2} (\sigma_{1,i}^{-2} - \sigma_{0,i}^{-2}) 
    \end{split}
\end{equation}

\section{Relationship between negative log-likelihood and Kullback-Leibler divergence} \label{app:nllferelation}

In this work, we trained our architecture with the negative log-likelihood as the loss but performed goal-directed control via EFE minimization which includes the Kullback-Leibler divergence.
Here, we show the relationship between the negative log-likelihood and the Kullback-Leibler divergence in general.

The Kullback-Leibler divergence between two probability distributions $p$ and $q$ is defined as
\begin{equation}
    \begin{split}
        D[p(x) \mid \mid q(x)] & = E_{x \sim p(x)} \bigg[ \log \frac{p(x)}{q(x)} \bigg] \\
        & = E_{x \sim p(x)} [\log p(x) - \log q(x)] \\
        & = E_{x \sim p(x)} [\log p(x)] - E_{x \sim p(x)} [\log q(x)]
    \end{split}
\end{equation}
where $E$ denotes the expected value.
Let us now assume that $P$ describes the distribution of some data we want to approximate with $q$.
The left term does not depend on $q$ and therefore is constant.
If we now take $N$ samples from the real distribution with $\lim_{N \to \infty}$ we end up with
\begin{equation}
    \begin{split}
        -E_{x \sim p(x)} [\log q(x)] & = - \frac{1}{N} \sum_{i}^N \log q(x) \\
    \end{split}
\end{equation}
which, up to a constant factor, is the definition of the negative log-likelihood.

We conclude that minimizing negative log-likelihood is equivalent to minimizing the Kullback-Leibler divergence.

\end{document}